\documentclass[10pt]{article}
\usepackage{geometry}
\usepackage{tikz}
\usetikzlibrary{shapes.geometric, arrows.meta, positioning, fit, backgrounds, calc, decorations.pathreplacing, shapes.misc}

\newcommand{\cmark}{\checkmark}  
\newcommand{\xmark}{--}          
\usepackage[table]{xcolor}
\usepackage{float}

\usepackage[utf8]{inputenc} 
\usepackage{hyperref}       
\usepackage{url}            
\usepackage{booktabs}       
\usepackage{amsfonts}       
\usepackage{nicefrac}       
\usepackage{microtype}      
\usepackage{lipsum}
\usepackage{natbib}
\usepackage{graphicx}
\usepackage{amsmath,amsfonts,amssymb}

\newtheorem{definition}{Definition}[section]  
\usepackage{threeparttable}

\usepackage{authblk} 
\usepackage{bbm}
\usepackage{algorithm}
\usepackage{algpseudocode}

\usepackage[utf8]{inputenc}
\usepackage{longtable}
\usepackage{multirow}
\usepackage{xcolor}

\title{FSL-BDP: Federated Survival Learning with Bayesian Differential Privacy for Credit Risk Modeling}

\author[1]{Sultan Amed \thanks{Corresponding author. Email: ef21sultanamed@iimidr.ac.in}}
\author[2]{Tanmay Sen}
\author[1]{Sayantan Banerjee}

\affil[1]{\small OM \& QT Area, Indian Institute of Management Indore, M.P. 453556, India.
}
\affil[2]{\small SQC \& OR Unit, Indian Statistical Institute Kolkata, W.B. 700108, India}

\date{}

\begin{document}
\maketitle

\begin{abstract}
Credit risk models are a critical decision-support tool for financial institutions, yet tightening data-protection rules (e.g., GDPR, CCPA) increasingly prohibit cross-border sharing of borrower data, even as these models benefit from cross-institution learning. Traditional default prediction suffers from two limitations: binary classification ignores default timing, treating early defaulters (high loss) equivalently to late defaulters (low loss), and centralized training violates emerging regulatory constraints. We propose a Federated Survival Learning framework with Bayesian Differential Privacy (FSL-BDP) that models time-to-default trajectories without centralizing sensitive data. The framework provides Bayesian (data-dependent) differential privacy (DP) guarantees while enabling institutions to jointly learn risk dynamics. Experiments on three real-world credit datasets (LendingClub, SBA, Bondora) show that federation fundamentally alters the relative effectiveness of privacy mechanisms. While classical DP performs better than Bayesian DP in centralized settings, the latter benefits substantially more from federation (+7.0\% vs +1.4\%), achieving near parity of non-private performance and outperforming classical DP in the majority of participating clients. This ranking reversal yields a key decision-support insight: privacy mechanism selection should be evaluated in the target deployment architecture, rather than centralized benchmarks. These findings provide actionable guidance for practitioners designing privacy-preserving decision support systems in regulated, multi-institutional environments.

\end{abstract}


\sloppy

\section{Introduction}
Rapid and sustained growth in digital lending has transformed global credit markets, expanding speed, accessibility, and financial inclusion to populations that were historically underserved. Recent data from the World Bank's Global Findex 2025 indicates that 79\% of adults globally held a financial account in 2024, up from 51\% in 2011~\cite{worldbank2025findex}. This expansion has been largely driven by digital platforms that lower barriers for unbanked and underbanked consumers, a trend further supported by mobile and fintech innovations in regions such as Sub-Saharan Africa and South Asia, where mobile money has substantially increased women's access to financial tools~\citep{klapper2024digital,worldbank2024ssa}. Peer-to-peer platforms such as LendingClub~\cite{zhang2020credit, serrano2016use, jagtiani2019roles, lyocsa2022default} in the United States and Bondora~\cite{lyocsa2022default, domotor2023peer} in Europe illustrate how low-cost, data-driven systems are reshaping retail lending, improving financial inclusion while introducing new challenges for risk management. The rise in granular data collection and application of artificial intelligence has further accelerated this shift, with credit scoring remaining central to financial decision-making~\cite{kowsar2023credit}. Credit scoring models serve as the core decision support artifact in lending operations, directly influencing approval rates, pricing strategies, portfolio risk management, and regulatory capital calculations. As such, advances in credit risk modeling constitute contributions to the decision support systems domain with immediate practical relevance.

\begin{figure}
    \centering
    \includegraphics[width=1.0\linewidth]{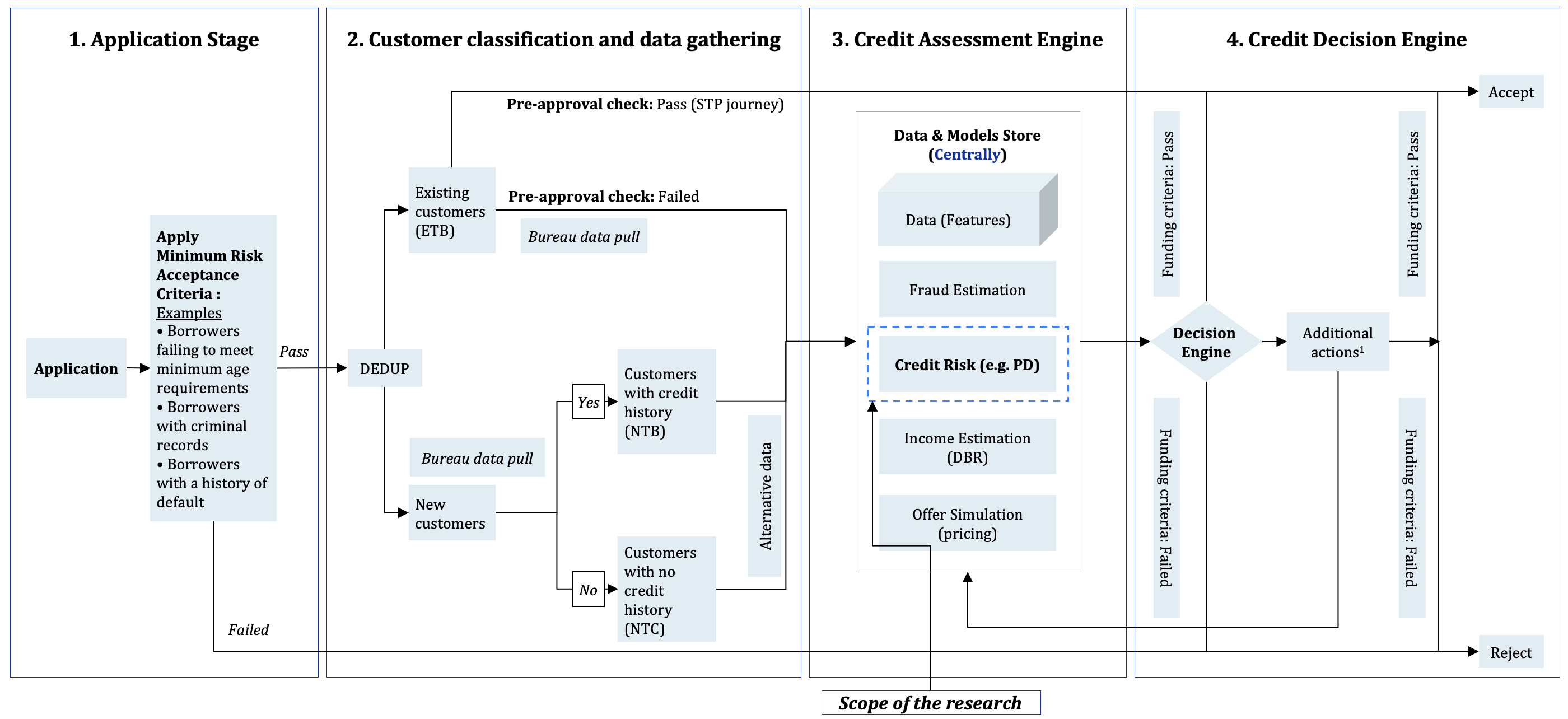}
    \caption{AI driven end-to-end credit decision engine}
    \label{fig:credit_engine}
\end{figure}

Digital lending relies on automated credit decision engines. As illustrated in Figure~\ref{fig:credit_engine}, the credit assessment process begins with basic eligibility checks including minimum acceptance criteria (e.g., age, prior default history, jurisdictional requirements). The system then performs deduplication to determine whether the applicant is an existing customer or a new applicant. Existing customers with strong internal credit scores may be processed via straight-through processing, while others undergo further assessment including fraud screening, income estimation, and computation of credit risk measures such as probability of default (PD), loss given default (LGD), and exposure at default (EAD). Among these, PD models exert the greatest influence on loan approvals, credit risk management, and portfolio monitoring, ultimately affecting financial stability~\cite{acemoglu2015systemic}. Accurate default risk assessment supports both institutional soundness and systemic resilience~\cite{khurram2025systemic,baesens2023boosting}.

\begin{figure}
        \centering
        \includegraphics[width=1\linewidth]{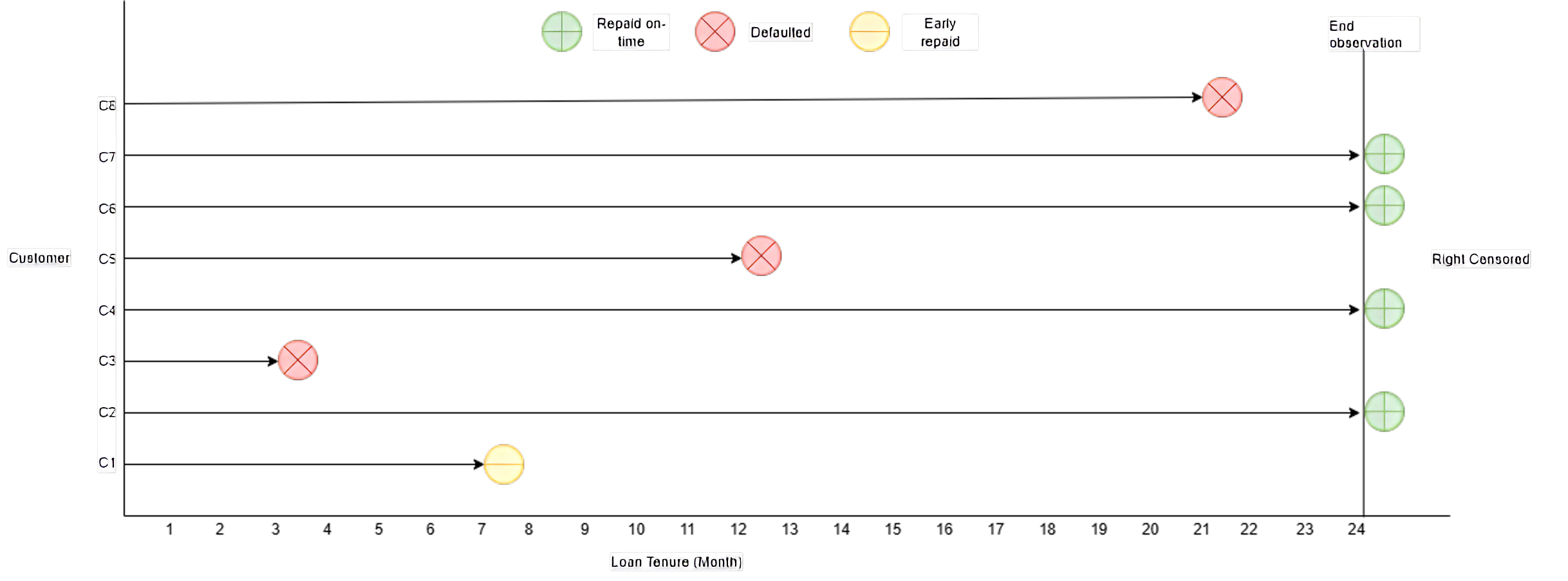}
        \caption{Time-to-event over loan tenure}
        \label{fig:survival_loan_tenure}
\end{figure}

Historically, PD scoring has been dominated by scorecards and logistic regression, valued for interpretability and regulatory acceptance~\cite{baesens2014analytics, baesens2023boosting}. Modern systems increasingly adopt machine learning and deep learning to improve predictive accuracy and capture nonlinear pattern~\cite{baesens2023boosting,zhao2015investigation,mahbobi2023credit}. Recent studies also demonstrate that alternative data sources such as telecom usage, utility payments, and online spending strengthen credit risk models, particularly for thin-file borrowers~\cite{djeundje2021enhancing,cornelli2023fintech}. Despite such progress, classical PD models face structural limitations. Most are framed as binary classification~\cite{koutanaei2015hybrid, fernandes2016spatial, croux2020important}, where target variables are defined within a fixed observation horizon (e.g., 90 days past due within 12 months~\cite{choy2011markov, brezigar2021modeling}). Both academic research and industry practice typically ignore default timing, failing to distinguish early from late defaulters. For example, as shown in Figure~\ref{fig:survival_loan_tenure}, a rule that defines default as “90 DPD ever in 12 months” would label \textcolor{blue}{C3} and \textcolor{blue}{C5} as defaults but ignore \textcolor{blue}{C8}, where the borrower defaults after the 12-month window. Early repayment cases like \textcolor{blue}{C1} are also commonly excluded from training, which removes useful behavioral signals. Current approaches fail to account for profitability because they treat \textcolor{blue}{C3} and \textcolor{blue}{C5} as equivalent despite meaningful differences in default timing. This issue is more pronounced in small-ticket, short-tenor loans, where LGD and EAD models are rarely used. These weaknesses reduce both decision accuracy and economic interpretability \citep{bellotti2009support,dirick2017time}. To address these challenges, this paper investigates two research questions:
\textcolor{blue}{(Q1)} How can decision support systems distinguish early defaulters from late defaulters to enable risk-adjusted pricing and provisioning decisions that reflect differential loss severity across loan tenure? and \textcolor{blue}{(Q2)} How can we model defaults beyond the traditional observation 
window, overcoming the truncation limitations of classification-based PD models?

Additionally, data protection regulations are reshaping the lending landscape. Privacy, data-residency, and responsible-AI rules increasingly limit cross-border and cross-institutional movement of borrower data. Frameworks such as GDPR in Europe~\cite{horobets2025artificial}, CCPA in the United States~\cite{varma2025federated}, and supervisory guidance from authorities including the EBA and MAS~\cite{tan2023data} constrain the central storage of sensitive attributes. As a result, consolidating data at a central hub is increasingly infeasible, even when larger pooled datasets would improve model quality. This creates fundamental challenges for financial institutions that require richer, more accurate models while complying with strict privacy and security requirements. Hence, we investigate two additional research questions: \textcolor{blue}{(Q3)} How can federated architectures enable collaborative decision support across institutions by learning repayment patterns across multiple geographies without exchanging raw data, while maintaining privacy and data security? and \textcolor{blue}{(Q4)} Under realistic non-IID client distributions and privacy controls, can a federated survival approach meet predefined performance and calibration thresholds, without unacceptable degradation relative to a centralized benchmark?

Federated learning offers a promising response to the data-sharing constraints, as it trains models where data reside and aggregates only model updates rather than raw observations~\cite{mcmahan2017communication}. Building on this rationale, we propose a Federated Survival Learning (FSL) framework that models time-to-default and trains collaboratively across institutions while keeping raw data local. The survival formulation preserves timing information and handles censoring naturally. The federated architecture enables cross-silo learning that respects data-residency constraints. We further strengthen the framework with Bayesian differential privacy mechanisms to bound information leakage under data-dependent privacy guarantees.

Despite progress in federated learning for classification-based credit scoring~\cite{shingi2020federated, wang2024novel, li2024dynamic}, no prior work has addressed federated survival analysis with formal privacy guarantees for time-to-default modeling. This paper bridges three research streams: (i) credit scoring and ML-based decision support, (ii) survival analysis for risk modeling, and (iii) federated and privacy-preserving machine learning. The key contributions are as follows: 
\begin{itemize}
     
    \item We adopt a survival-based credit risk formulation that distinguishes early and late defaulters, allowing the model to capture how default risk evolves over time.

    \item We develop a federated survival learning framework, FSL-BDP, that enables collaborative model training across multiple financial institutions and geographic regions while providing Bayesian privacy guarantees through Bayesian Differential Privacy (BDP). The framework perturbs client-level gradient information using a Bayesian privacy accounting mechanism, thereby reducing inferential privacy risks such as membership and attribute inference attacks. FSL-BDP is explicitly designed to operate under realistic non-IID data distributions that arise naturally in cross-institutional credit datasets. To the best of our knowledge, this study represents the first application of Bayesian Differential Privacy within a Federated survival analysis setting.

    \item We evaluate FSL-BDP on three real-world credit risk datasets, namely, LendingClub, SBA, and Bondora, using both in-time (ITV) and out-of-time (OOT) validation under federated, privacy-constrained, and natural non-IID settings. The results show that BDP provides a better privacy-utility trade-off than classical DP and fundamentally alters the relative effectiveness of privacy mechanisms. This indicates that privacy mechanism selection should be evaluated in the target deployment architecture rather than on centralized benchmarks.
    
\end{itemize}  

We emphasize that our privacy guarantees are based on Bayesian Differential Privacy, which provides distribution-dependent protection rather than worst-case guarantees. While this distinction is important for regulatory interpretation, BDP has been shown to offer substantially improved utility in complex FL settings.

The remainder of this paper is organized as follows. Section~\ref{sec:related_work} reviews related work on credit scoring, survival analysis, federated learning, and privacy-preserving methods. Section~\ref{sec:Methods} details the proposed methodology, including the Federated Survival Learning framework and optimization techniques. Section~\ref{sec:EXPERIMENT_SETUP} describes the experimental design and datasets. Section~\ref{sec:Results_Discussions} reports the empirical results, while Section~\ref{sec:Discussion} discusses the findings, managerial implications, and directions for future research. We finally present the conclusions in Section~\ref{sec:Conclusion}.

\section{Related Work}
\label{sec:related_work}

The rapid growth of digital lending has been extensively 
documented~\cite{cornelli2023fintech, suryono2019peer}, alongside the continued expansion of data-driven decision-making and AI in lending~\cite{kowsar2023credit, sadok2022artificial, cao2020ai}. Credit scoring has been central to lending decisions for decades~\cite{baesens2023boosting, baesens2014analytics, klimowicz2021concept} and has evolved significantly over the past three to four decades. The field began with expert-assigned scorecards \citep{clauser1995scoring,mays1995handbook}, progressed through traditional statistical models such as logistic regression, and advanced to tree-based ensembles machine learning \citep{li2020comparative,wang2011comparative} (e.g., AdaBoost, XGBoost \citep{qin2021xgboost}, LightGBM \citep{lextrait2023scaling}, random forests \citep{zhang2018novel,arora2020bolasso}) and deep learning \citep{zhang2020deep} to capture high-dimensional, nonlinear borrower signals. Substantial recent literature suggests that predictive lift now more 
often arises from stronger feature engineering~\cite{stevenson2021value, boughaci2018new, koutanaei2015hybrid}, calibrated probability estimates, 
and rigorous validation rather than from model complexity alone. Regulators and risk functions increasingly expect transparent reasoning artifacts such as SHAP and LIME explanations~\cite{chen2024interpretable}, alongside fairness considerations~\cite{moldovan2023algorithmic, kozodoi2022fairness} and accuracy measures.  Surveys and empirical work also document growing reliance on alternative data such as rent and utilities, telco, and e-commerce spend to score thin-file borrowers and promote inclusion \citep{djeundje2021enhancing,oskarsdottir2019value,fernandes2016spatial}, while cautioning about drift, disparate impact, and operational risks, including data privacy, in large-scale deployment \cite{amed2025pdx}. A common pattern in prior credit scoring research is framing default prediction as a binary classification task~\cite{lessmann2015benchmarking, moscato2021benchmark}, estimating the likelihood of default (e.g., 30+/60+/90+ days past due) within a fixed observation window. This formulation overlooks the time-to-default dimension that lenders need to distinguish early from late defaulters, motivating survival-based approaches.

\begin{table}[h]
\centering
\caption{Summary of related work in credit scoring, survival analysis, and federated learning}
\label{tab:related_work}
\footnotesize
\begin{tabular}{@{}lcccccc@{}}
\toprule
\textbf{Study} & \textbf{Class.} & \textbf{Surv.} & \textbf{Fed.} & \textbf{Privacy} & \textbf{Non-IID} & \textbf{Domain} \\
\midrule
\multicolumn{7}{@{}l}{\textit{Credit Scoring (Classification)}} \\
Hsieh et al. (2010) \cite{hsieh2010data} & \cmark & \xmark & \xmark & \xmark & \xmark & Credit \\
Lessmann et al. (2015) \cite{lessmann2015benchmarking} & \cmark & \xmark & \xmark & \xmark & \xmark & Credit \\
Papouskova et al. (2019) \cite{papouskova2019two} & \cmark & \xmark & \xmark & \xmark & \xmark & Credit \\
Gunnarsson et al. (2021) \cite{gunnarsson2021deep} & \cmark & \xmark & \xmark & \xmark & \xmark & Credit \\
\midrule
\multicolumn{7}{@{}l}{\textit{Survival Analysis (Centralized)}} \\
Dirick et al. (2017) \cite{dirick2017time} & \xmark & \cmark & \xmark & \xmark & \xmark & Credit \\
Xia et al. (2021) \cite{xia2021dynamic} & \xmark & \cmark & \xmark & \xmark & \xmark & Credit \\
Bai et al. (2022) \cite{bai2022gradient} & \xmark & \cmark & \xmark & \xmark & \xmark & Credit \\
Blumenstock et al. (2022) \cite{blumenstock2022deep} & \xmark & \cmark & \xmark & \xmark & \xmark & Credit \\
Medina-Olivares et al. (2023) \cite{medina2023joint} & \xmark & \cmark & \xmark & \xmark & \xmark & Credit \\
\midrule 
\multicolumn{7}{@{}l}{\textit{Federated Credit Scoring}} \\
Shingi et al. (2020) \cite{shingi2020federated} & \cmark & \xmark & \cmark & \xmark & \xmark & Credit \\
He et al. (2023) \cite{he2023privacy} & \cmark & \xmark & \cmark & \xmark & \xmark & Credit \\
Wang et al. (2024) \cite{wang2024novel} & \cmark & \xmark & \cmark & \xmark & \xmark & Credit \\
Li et al. (2024) \cite{li2024dynamic} & \cmark & \xmark & \cmark & \xmark & \xmark & Credit \\
\midrule
\multicolumn{7}{@{}l}{\textit{Federated Survival Learning}} \\
Andreux et al. (2020) \cite{andreux2020federated} & \xmark & \cmark & \cmark & \xmark & \xmark & Health \\
Rahman et al. (2023) \cite{rahman2023fedpseudo} & \xmark & \cmark & \cmark & \xmark & \cmark & Health \\
Archetti et al. (2023) \cite{archetti2023federated} & \xmark & \cmark & \cmark & \xmark & \xmark & Health \\
Wen et al.  (2025) \cite{wen2025federated} & \xmark & \cmark & \cmark & Classical & \xmark & Health \\
Veeraragavan et al. (2025) \cite{veeraragavan2024differentially} & \xmark & \cmark & \cmark & Classical & \xmark & Health \\
\midrule
\textbf{This work (FSL-BDP)} & \xmark & \cmark & \cmark & \textbf{BDP} & \cmark & \textbf{Credit} \\
\bottomrule
\end{tabular}

\vspace{0.8em}
\raggedright
\footnotesize
\textit{Notes}: Class. = Classification; Surv. = Survival analysis; Fed. = Federated learning; BDP = Bayesian differential privacy. $\checkmark$ = work present; -- = work absent. To our knowledge, this work is the first to combine federated survival analysis with Bayesian differential privacy for credit risk modeling.
\end{table}

Survival analysis has been extensively used in medicine since the middle of the last century, with major advances including the Kaplan–Meier estimator and Cox’s proportional hazards model~\cite{kaplan1958nonparametric, cox1972regression}. Narain~\cite{narain1992survival} first introduced survival models to credit risk, highlighting their ability to estimate not only whether but also when a borrower will default. Subsequent work applied Cox proportional hazards and accelerated failure time models to capture borrower-specific risk over time~\cite{dirick2017time,hassan2018modeling}. More flexible specifications include spline-based hazards for nonlinear effects~\cite{luo2016spline} and mixture-cure models that distinguish likely defaulters from borrowers who are unlikely to default within the loan horizon~\cite{tong2012mixture}. Recent studies further advanced performance using boosted tree-based survival methods~\cite{xia2021dynamic, bai2022gradient} and neural survival networks for time-to-default modeling~\cite{blumenstock2022deep}. Joint models combining longitudinal behavioral data with survival outcomes have also been proposed~\cite{medina2023joint}. Despite these advantages, survival analysis is still less commonly adopted in production credit scoring systems, and most existing implementations assume centralized training, raising data-privacy concerns under tightening regulatory requirements.

The foundational work on federated learning (FL) originated at Google, where models were trained locally on mobile devices and aggregated using Federated Averaging (FedAvg)~\cite{mcmahan2017communication,hard2018federated}. Since then, FL has emerged as a prominent paradigm in privacy-sensitive domains such as healthcare and finance, enabling multi-institution collaboration without centralizing raw data. In credit risk, Shingi~\cite{shingi2020federated} demonstrated federated logistic regression for credit scoring, while Wang et al.~\cite{wang2024novel} proposed federated knowledge transfer combining fine-tuning and knowledge distillation. Li et al.~\cite{li2024dynamic} introduced dynamic receptive fields for federated credit assessment. As summarized in Table~\ref{tab:related_work}, existing federated learning research in credit risk is largely confined to classification-based probability-of-default models and typically lacks formal privacy guarantees, while survival-based and privacy-aware federated approaches have seen limited adoption in financial risk modeling, leaving time-to-default learning under realistic non-IID financial data largely unexplored.

Differential privacy (DP) has become the standard for providing formal privacy guarantees in machine learning, with DP-SGD~\cite{abadi2016deep} offering a principled mechanism for private gradient-based optimization. However, DP-SGD introduces a substantial privacy–utility trade-off: worst-case noise calibration required to satisfy $(\varepsilon, \delta)$-DP can significantly degrade model performance, particularly under strict privacy budgets~\cite{wei2020federated, triastcyn2019federated}. Bayesian differential privacy (BDP) has been proposed to address this limitation by leveraging data-dependent privacy accounting rather than worst-case bounds, enabling tighter privacy guarantees with reduced noise injection~\cite{triastcyn2019federated, triastcyn2020bayesian}. Triastcyn and Faltings~\cite{triastcyn2019federated} demonstrated the benefits of BDP in federated image classification, showing improved utility retention relative to classical DP-SGD. However, BDP has thus far been studied primarily in classification tasks on image data, and its application to survival analysis, particularly for time-to-default modeling in credit risk, has not been examined.

Our work addresses this gap by integrating Bayesian differential privacy with federated survival learning, employing leave-one-out gradient estimation and pairwise Rényi divergence to achieve strong privacy guarantees while preserving predictive utility. To our knowledge, this is the first study to apply Bayesian differential privacy to survival analysis, and the first to do so in the context of federated credit risk modeling.

\section{Methodologies}

\label{sec:Methods}
This section presents the proposed federated survival analysis framework with Bayesian differential privacy (BDP) guarantees. We begin with centralized survival modeling as a baseline, then progressively introduce federated learning and privacy-preserving mechanisms, culminating in our novel Bayesian DP approach to model time-to-default for credit risk assessment.

\subsection{Problem Formulation}
\label{subsec:problem}

Consider a federated credit risk setting  with $K$ participating institutions (clients), where the $k$-th client possesses a local dataset $\mathcal{D}_k = \{(\mathbf{x}_i^{(k)}, t_i^{(k)}, \delta_i^{(k)})\}_{i=1}^{n_k}$, where $\mathbf{x}_i^{(k)} \in \mathbb{R}^d$ denotes the $d$-dimensional borrower covariate vector at client $k$, $t_i^{(k)} \in \mathbb{R}^+$ represents the observed time-to-default or censoring, and $\delta_i^{(k)} \in \{0,1\}$ is the event indicator, where $\delta_i^{(k)} = 1$ indicates default occurrence and $\delta_i^{(k)} = 0$ denotes right censoring.

The global dataset comprises $N = \sum_{k=1}^K n_k$ observations, where $n_k$ is number of observations at client k. The objective is to learn a global survival model $f_\theta: \mathbb{R}^d \to [0,1]^T$ that predicts discrete-time survival probabilities at $T$ time intervals, while providing formal differential privacy guarantees for individual-level data.

We establish  baseline using discrete-time survival analysis with neural networks, following the framework of \citep{kvamme2019time,gensheimer2019scalable} for both centralized and federated learning.

\subsection{Discrete-Time Survival Likelihood}
\label{subsec:survival_likelihood}

Since loan repayments occur monthly, we model time in discrete form using $T$ ordered, left-closed and right-open intervals,
\[ 0 = \tau_0 < \tau_1 < \cdots < \tau_T < \infty. \]

Let $h_l^{(k)}(\mathbf{x})$ denote the discrete-time hazard probability for a borrower
at client $k$ in interval $(\tau_{l-1}, \tau_l]$, defined as
\begin{equation*}
h_l^{(k)}(\mathbf{x})
=
P\!\left(
t \in (\tau_{l-1}, \tau_l]
\mid
t > \tau_{l-1}, \mathbf{x}, k
\right),
\quad l = 1,\ldots,T.
\label{eq:discrete_hazard}
\end{equation*}

This formulation does not impose the proportional hazards assumption and allows for flexible,
time-varying default dynamics. The corresponding survival function up to the end of interval $l$ is
\begin{equation*}
S_l^{(k)}(\mathbf{x})
=
P(t > \tau_l \mid \mathbf{x})
=
\prod_{j=1}^{l} \big(1 - h_j^{(k)}(\mathbf{x})\big).
\end{equation*}

\paragraph{Individual Likelihood: } Let $j$ denote the interval such that $t \in (\tau_{j-1}, \tau_j]$. If an uncensored individual experiences the event in interval $j$, the likelihood is
\begin{equation*}
\mathcal{L}
=
h_j^{(k)}(\mathbf{x})
\prod_{l=1}^{j-1} \big(1 - h_l^{(k)}(\mathbf{x})\big).
\end{equation*}

For a right-censored individual with censoring occurring in interval $j$, the likelihood is
\begin{equation*}
\mathcal{L}
=
\prod_{l=1}^{j} \big(1 - h_l^{(k)}(\mathbf{x})\big).
\end{equation*}

\paragraph{Likelihood Representation:}
For computational convenience, each individual is represented using two binary vectors:
$\mathbf{s}^{\text{surv}} \in \{0,1\}^T$ and $\mathbf{s}^{\text{fail}} \in \{0,1\}^T$.
The vector $\mathbf{s}^{\text{surv}}$ indicates the intervals through which the individual has survived, and $\mathbf{s}^{\text{fail}}$ indicates the interval in which failure occurs.

For an uncensored individual with event time $t \in (\tau_{l-1}, \tau_l]$,
\[
s^{\text{surv}}_l =
\begin{cases}
1, & t \geq \tau_l, \\
0, & \text{otherwise},
\end{cases}
\qquad
s^{\text{fail}}_l =
\begin{cases}
1, & \tau_{l-1}\leq t < \tau_l, \\
0, & \text{otherwise}.
\end{cases}
\]

For a right-censored individual with censoring time $t$, the survival indicator vector
$\mathbf{s}^{\text{surv}}$ is defined using the midpoint convention:
\[
s^{\text{surv}}_l =
\begin{cases}
1, & \text{if } t \geq \dfrac{\tau_{i-1} + \tau_i}{2}, \\
0, & \text{otherwise},
\end{cases}
\qquad
s^{\text{fail}}_l = 0 .
\]

Using this representation, the individual log-likelihood can be written compactly as
\begin{equation}
\log \mathcal{L}
=
\sum_{l=1}^{T}
\left[
s^{\text{surv}}_l \log \big(1 - h_l^{(k)}(\mathbf{x})\big)
+
s^{\text{fail}}_l \log h_l^{(k)}(\mathbf{x})
\right].
\label{eq:vectorized_loglik}
\end{equation}

The total loss is obtained by summing the negative log-likelihood in
Eq.~\eqref{eq:vectorized_loglik} over all individuals and is minimized using stochastic or mini-batch
gradient descent.

\paragraph{Neural Network Architecture: }
We parameterize the client-specific discrete-time hazard function using a neural network:
\begin{equation}
h_l^{(k)}(\mathbf{x}; \theta)
=
\sigma\!\left(f_l(\mathbf{x}; \theta)\right),
\quad l = 1,\ldots,T,
\end{equation}
where $\sigma(\cdot)$ denotes the sigmoid activation function, ensuring
$h_l^{(k)}(\mathbf{x}; \theta) \in (0,1)$.

Specifically, we employ a feed-forward neural network whose output layer has $T$ units,
each corresponding to one discrete time interval:
\begin{equation}
\mathbf{h}^{(k)}(\mathbf{x}; \theta)
= \sigma\!\left(\mathbf{W}_L \,\phi_{L-1}\!\left(\mathbf{W}_{L-1} \,\phi_{L2}\!\left(\cdots \phi_1(\mathbf{W}_1 \mathbf{x} + \mathbf{b}_1)\cdots
\right)+ \mathbf{b}_{L-1}\right)+ \mathbf{b}_L\right),
\label{eq:nn_architecture}
\end{equation}

where $\theta = \{\mathbf{W}_l, \mathbf{b}_l\}_{l=1}^{L}$ denotes the shared network parameters,
$\phi_l(\cdot)$ is the activation function of the $l$-th hidden layer (e.g., SELU),
and $\sigma(\cdot)$ is applied element-wise to ensure valid hazard probabilities.
The $l$-th output node corresponds to the conditional hazard probability
$h_l^{(k)}(\mathbf{x}; \theta)$ for interval $(\tau_{l-1}, \tau_l]$.

\begin{figure}[H]
\label{fig_fl}
    \centering
    \includegraphics[width=0.9\linewidth]{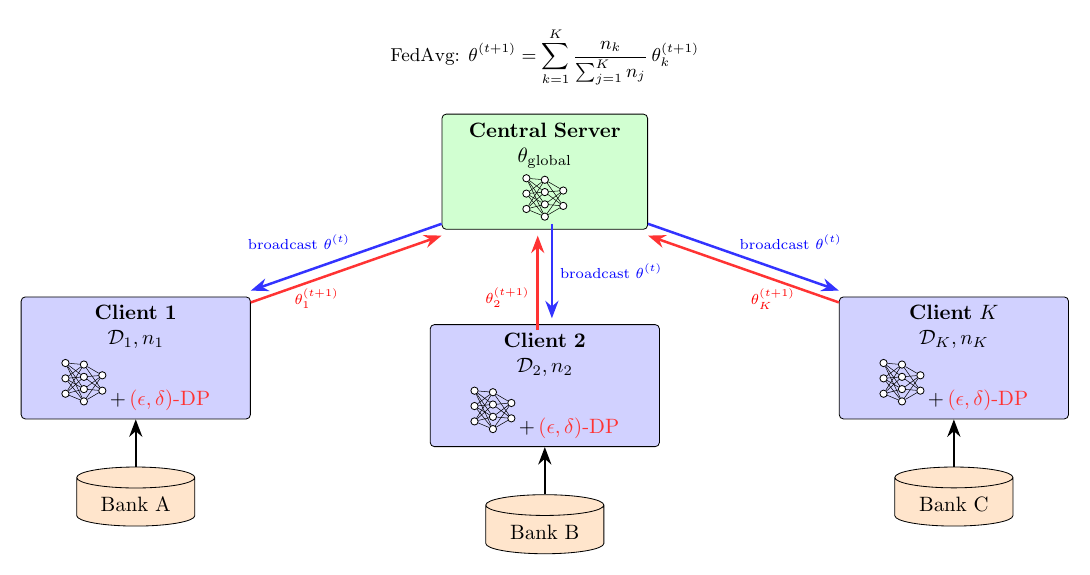}
    \caption{Federated Differential Privacy}
\end{figure}

\subsection{Federated Learning}
Traditional centralized machine learning requires aggregating all client data on a single server, which is often infeasible in credit risk modeling due to strict privacy, security, and regulatory constraints. Federated Learning (FL), introduced by \cite{mcmahan2017communication, hard2018federated}, addresses these challenges by enabling collaborative model training without sharing raw data. In FL, each client locally trains a survival model on its private credit portfolio and transmits only model updates to a central server. The server aggregates these updates typically via weighted averaging to obtain a global model, which is then redistributed to clients for the next training round. This iterative process preserves data locality while enabling collective learning across distributed institutions. The standard FL architecture with differential privacy is illustrated in Figure~\ref{fig_fl}.

In real-world credit risk modeling, borrower populations, default dynamics, and censoring mechanisms vary substantially across financial institutions and geographic regions. As a result, the client-level datasets
$\{\mathcal{D}_k\}_{k=1}^{K}$ are generally \emph{non-independently and identically distributed (non-IID)}. We assume that, for each client $k$, the observed triples
$(\mathbf{x}_i^{(k)}, t_i^{(k)}, \delta_i^{(k)})$, $i=1,\ldots,n_k$,
are independently drawn from a client-specific joint distribution
\[
(\mathbf{x}^{(k)}, t^{(k)}, \delta^{(k)}) \sim \mathcal{P}_k,
\]
where, in general,
\[
\mathcal{P}_k \neq \mathcal{P}_j \quad \text{for } k \neq j.
\]

Under this setting, the local survival risk at client $k$ is defined as
\[
\mathcal{L}_k(\theta)
=
\mathbb{E}_{(\mathbf{x}^{(k)}, t^{(k)}, \delta^{(k)})\sim\mathcal{P}_k}
\left[
\ell\big(\mathbf{x}^{(k)}, t^{(k)}, \delta^{(k)}; \theta\big)
\right],
\]
while the global federated objective minimizes a mixture risk
\[
\mathcal{L}(\theta)
=
\sum_{k=1}^{K} \pi_k \mathcal{L}_k(\theta),
\quad
\pi_k = \frac{n_k}{\sum_{j=1}^{K} n_j}.
\]

This formulation explicitly acknowledges that federated survival learning optimizes a shared model across heterogeneous credit populations, rather than assuming identical data-generating processes across clients.

\subsection{Differential Privacy}

Although Federated Learning keeps raw credit data localized at client institutions, it does not fully eliminate privacy risks. In particular, sensitive information may still be inferred from the model updates transmitted to the central server, for example, through gradient leakage or membership inference attacks \cite{dwork2014algorithmic, wei2020federated}. Therefore, formal privacy guarantees are required when training federated survival models on sensitive time-to-event credit data. Differential Privacy provides such guarantees by quantifying how much the output of a learning algorithm changes when a single data point in the training dataset is modified \cite{abadi2016deep}. 

\begin{definition}[($\epsilon, \delta$)-Differential Privacy \cite{abadi2016deep}]
A randomized mechanism $\mathcal{M}: \mathcal{D} \to \mathcal{R}$ satisfies $(\epsilon, \delta)$-differential privacy if for all neighboring datasets $\mathcal{D}, \mathcal{D}'$ differing in one record, and for all measurable sets $S \subseteq \mathcal{R}$:
\begin{equation}
P(\mathcal{M}(\mathcal{D}) \in S) \leq e^\epsilon \cdot P(\mathcal{M}(\mathcal{D}') \in S) + \delta.
\label{eq:dp_definition}
\end{equation}
\end{definition}

Here, $\epsilon > 0$ controls the strength of the privacy guarantee, with smaller values implying stronger privacy protection, while $\delta \geq 0$ denotes the probability of privacy failure and is typically required to be negligible, often less than $1/|\mathcal{D}|$ \cite{dwork2014algorithmic}. Setting $\delta = 0$ recovers pure $\epsilon$-differential privacy.

To enforce DP in deep neural networks, including survival models trained in federated settings, DP-SGD \cite{abadi2016deep} modifies standard stochastic gradient descent by clipping per-sample gradients and injecting Gaussian noise. Specifically, for a mini-batch of size $L$, the sanitized gradient at iteration $r$ is computed as
\begin{equation}
\mathbf{\tilde{g_r}(x_i)} = \frac{1}{L}\Bigg( \sum_{i=1}^L\frac{\mathbf{g_r(x_i)}}{\max\left(1, \frac{||\mathbf{g_r(x_i)}||_2}{C}\right)} + \mathcal{N}(0, \sigma^2 C^2 \mathbf{I})\Bigg), ~i = 1,2,\cdots,L,
\end{equation}

where $C$ is the gradient clipping threshold and $\sigma$ controls the noise magnitude. Each training iteration thus satisfies differential privacy, and the overall privacy budget across $R$ federated rounds is obtained via composition.

\paragraph{Privacy Accounting via Moments Accountant:}
\label{subsec:privacy_accounting}

The privacy loss incurred during training accumulates across optimization steps.
Let $R$ denote the total number of optimization steps at which differential privacy is enforced
(e.g., local DP-SGD updates across federated training rounds), and let
$q = B / n_k$ be the subsampling rate, where $B$ is the mini-batch size and $n_k$ is the
dataset size at a given client.
Following the moments accountant framework \citep{abadi2016deep, geyer2017differentially},
which can be equivalently expressed using Rényi Differential Privacy (RDP)
\citep{mironov2017renyi}, we track the cumulative privacy loss throughout training.

\paragraph{Rényi Differential Privacy (RDP) \citep{mironov2017renyi}:}
Under Poisson subsampling with rate $q$, the Gaussian mechanism with noise scale $\sigma$
satisfies RDP at order $\alpha > 1$.
After $R$ compositions, the total RDP budget is bounded by
\begin{equation}
\epsilon_\alpha(R)
\leq
R \cdot \frac{\alpha q^2}{2\sigma^2}
+
O(R q^3),
\label{eq:rdp}
\end{equation}
where higher-order terms become negligible for small subsampling rates.
The final $(\epsilon,\delta)$-differential privacy guarantee is obtained via
\begin{equation}
\epsilon
=
\min_{\alpha > 1}
\left\{
\epsilon_\alpha(R)
+
\frac{\log(1/\delta)}{\alpha - 1}
\right\},
\label{eq:rdp_to_dp}
\end{equation}
which yields a tighter privacy bound than classical composition rules.
This privacy accounting mechanism forms the foundation upon which we build our Bayesian differential privacy (BDP) \cite{triastcyn2020bayesian} enhanced federated survival learning (FSL-BDP) framework for credit risk modeling.

\subsection{Bayesian Differential Privacy Framework}

Bayesian Differential Privacy (BDP) \cite{triastcyn2020bayesian} refines classical differential
privacy by measuring privacy loss in terms of posterior distributions over model parameters,
rather than worst-case differences in algorithm outputs.
Specifically, BDP quantifies how much an adversary’s posterior belief about the model parameters
can change after observing the output of a learning algorithm, given a prior belief and an
assumed data-generating distribution. 

Unlike standard differential privacy, which enforces uniform guarantees over all possible
datasets, BDP assumes that data points are drawn i.i.d.\ from an underlying distribution.
This enables tighter, data-dependent privacy guarantees that are particularly well suited
to complex learning models. As a consequence, $(\epsilon_\mu, \delta_\mu)$-BDP guarantees should be interpreted as average-case or high-probability bounds under the assumed data generating distribution, and are not equivalent to the worst-case $(\epsilon, \delta)$-DP guarantees.

To the best of our knowledge, BDP has not been applied to federated survival analysis,
where privacy is critical due to sensitive time-to-event data and heterogeneous client populations.

\begin{definition}[Bayesian Differential Privacy \cite{triastcyn2020bayesian}]
A randomized learning algorithm $\mathcal{A}$ satisfies
$(\epsilon_\mu, \delta_\mu)$-Bayesian Differential Privacy if for any two neighboring datasets
$\mathcal{D}$ and $\mathcal{D}'$ differing in one sample drawn from $\mu(x)$,
the privacy loss random variable
\[
L(\theta; \mathcal{D}, \mathcal{D}')
=
\log \frac{p(\theta \mid \mathcal{D})}{p(\theta \mid \mathcal{D}')}
\]
satisfies
\begin{equation}
\Pr_{\theta \sim \mathcal{A}(\mathcal{D})}
\left[
L(\theta; \mathcal{D}, \mathcal{D}') \le \epsilon_\mu
\right]
\ge 1 - \delta_\mu.
\label{eq:bayesian_dp}
\end{equation}
\end{definition}

This formulation aligns naturally with federated survival learning, where privacy guarantees
are enforced at the model-parameter level while accommodating censored time-to-event data
and heterogeneous client distributions. The i.i.d.\ assumption applies locally within each client,
while allowing client-specific data-generating processes across the federation.

\begin{algorithm}[H]
\caption{Federated Survival Learning with Bayesian Differential Privacy (FSL-BDP)}
\label{alg:bayesian_dp_federated}
\begin{algorithmic}[1]

\Require 
Number of clients $K$; client datasets $\{\mathcal{D}_k\}_{k=1}^K$; 
communication rounds $R$; local epochs $E$; client participation rate $p$; 
Bayesian DP parameters $(\sigma, C, \Lambda, \beta, \gamma)$

\Ensure 
Final global survival model $\theta_{\text{global}}^{(R)}$; 
per-client cumulative Bayesian DP guarantees $\{(\epsilon_k,\delta_k)\}_{k=1}^K$

\vspace{0.2em}
\State \textbf{Server:} Initialize global model $\theta_{\text{global}}^{(0)}$
\State Initialize cumulative privacy costs:
$\epsilon_k \gets 0,\ \delta_k \gets 0,\ \forall k$

\vspace{0.3em}
\For{$r = 0$ to $R-1$}

    \State \textbf{Server:} Sample participating clients
    \[
    \mathcal{S}_t \sim \mathrm{Uniform}\big(\{1,\ldots,K\}, \lfloor pK \rfloor\big)
    \]

    \State \textbf{Server:} Broadcast global model $\theta_{\text{global}}^{(r)}$ to all $k \in \mathcal{S}_r$

    \For{\textbf{each} client $k \in \mathcal{S}_t$ \textbf{in parallel}}

        \State Initialize local model:
        $\theta_k^{(r)} \gets \theta_{\text{global}}^{(r)}$

        \State
        \[
        \theta_k^{(r+1)},\,
        (\epsilon_k^{(r)}, \delta_k^{(r)})
        \gets
        \textsc{BayesianDP\_Train}
        \big(
            \theta_k^{(r)},
            \mathcal{D}_k,
            E,
            \sigma,
            C,
            \Lambda,
            \beta,
            \gamma
        \big)
        \]
        \Comment{Bayesian DP-SGD, Alg.~\ref{alg:bdp_sgd_mc}}

        \State Update cumulative privacy:
        \[
        \epsilon_k \gets \epsilon_k + \epsilon_k^{(r)},\quad
        \delta_k \gets \delta_k + \delta_k^{(r)}
        \]

        \State Upload privatized local model $\theta_k^{(r+1)}$ to server

    \EndFor

    \vspace{0.2em}
    \State \textbf{Server:} Aggregate received models (post-processing):
    \[
    \theta_{\text{global}}^{(r+1)} =
    \sum_{k \in \mathcal{S}_r}
    \frac{|\mathcal{D}_k|}{\sum_{j \in \mathcal{S}_r} |\mathcal{D}_j|}
    \,\theta_k^{(r+1)}
    \]

\EndFor

\vspace{0.2em}
\State \textbf{return}
$\theta_{\text{global}}^{(R)}$,
$\{(\epsilon_k,\delta_k)\}_{k=1}^K$

\end{algorithmic}
\end{algorithm}

\paragraph{Privacy Accounting: }
Following \citep{triastcyn2020bayesian}, privacy loss is quantified by comparing the
distributions of noisy model updates produced by neighboring datasets that differ
in a single data point drawn from the underlying data-generating distribution.
Rather than relying on worst-case sensitivity, Bayesian Differential Privacy estimates
privacy loss by evaluating the Rényi divergence between the induced distributions
of model parameters (or gradients) under data inclusion and exclusion.

In practice, this is implemented by approximating the distributions of per-sample
(or sub-sampled) gradients and computing the corresponding Bayesian Rényi divergence
at order $\lambda > 1$. Privacy loss is accumulated across local optimization steps
and federated communication rounds using Rényi composition. The accumulated Bayesian
Rényi privacy cost is then converted into an $(\epsilon_\mu, \delta_\mu)$ guarantee
using the Bayesian DP tail bound of \citep{triastcyn2020bayesian}, given by
\begin{equation}
\epsilon_\mu(\lambda)
=
\frac{C_{\mathrm{tot}}(\lambda) - \log \beta}{\lambda},
\qquad
\delta_\mu = \beta + \gamma,
\end{equation}
where $\beta$ denotes the privacy failure probability, $C_{\mathrm{tot}}(\lambda)$ denotes cumulative privacy cost,  and $\gamma$ accounts for
the Monte-Carlo estimation error. The final privacy level is obtained by minimizing
$\epsilon_\mu(\lambda)$ over $\lambda > 1$. The Bayesian Differential privacy algorithm is given in Algorithm \ref{alg:bdp_sgd_mc}.

Overall, Bayesian Differential Privacy enables tighter, data-aware privacy accounting
in federated learning and is particularly well suited for heterogeneous and non-IID
survival data.

\begin{algorithm}[!htbp]
\caption{Bayesian Differential Privacy: Bayesian DP-SGD}
\label{alg:bdp_sgd_mc}
\begin{algorithmic}[1]

\Require
Dataset $\mathcal{D}_k=\{x_i\}_{i=1}^{n_{k}}$;
data-generating prior $\mu(x)$;
initial parameters $\theta_0$;
learning rate $\eta$;
noise multiplier $\sigma$;
clipping norm $C$;
batch size $B$;
iterations $E$;
Rényi orders $\Lambda$;
Monte-Carlo samples $M$;
estimator failure probability $\gamma$;
privacy failure probability $\beta$

\Ensure
Final parameters $\theta_E$;
Bayesian DP guarantee $(\epsilon_\mu, \delta_\mu = \beta + \gamma)$

\vspace{0.3em}
\State Initialize model parameters $\theta \gets \theta_0$

\For{$\lambda \in \Lambda$}
    \State Initialize cumulative privacy cost $C_{\mathrm{tot}}(\lambda) \gets 0$
\EndFor

\vspace{0.3em}
\For{$e = 1$ to $E$}

    \State Sample mini-batch $B_e \subset D$ uniformly without replacement
    \State Compute per-sample gradients $\{ g_i(\theta) \}_{i \in B_e}$
    \State Clip gradients: $\bar g_i \gets g_i / \max\!\left(1, \frac{\|g_i\|_2}{C}\right)$
    \State Compute averaged gradient: $g_e \gets \frac{1}{|B_t|} \sum_{i \in B_e} \bar g_i$
    \State Sample noise $\xi_e \sim \mathcal{N}(0, \sigma^2 C^2 I)$
    \State Update parameters: $\theta \gets \theta - \eta (g_e + \xi_e)$
    \vspace{0.3em}
    \Statex \textcolor{blue}{// Monte-Carlo Bayesian Sensitivity Estimation}

    \For{$m = 1$ to $M$}
        \State Sample alternative data point $x'^{(m)} \sim \mu$
        \State Construct neighboring batch $B_e^{(m)}$ by replacing one element of $B_e$ with $x'^{(m)}$
        \State Compute clipped averaged gradient $g_e^{(m)}$
        \State Compute squared deviation: $\Delta_e^{(m)} \gets \| g_e - g_e^{(m)} \|_2^2$
    \EndFor

    \vspace{0.3em}
    \Statex \textcolor{blue}{// Bayesian Privacy Accounting}

    \State Set subsampling rate $q \gets \frac{B}{N}$

    \For{$\lambda \in \Lambda$}

        \For{$m = 1$ to $M$}
            \State Compute left privacy cost:$c_{e,Left}^{(m)}(\lambda)
            \gets
            \log \mathbb{E}_{K \sim \mathrm{Bin}(\lambda+1, q)}
            \left[
            \exp\!\left(
            \frac{K^2 - K}{2\sigma^2} \Delta_e^{(m)}
            \right)
            \right]$

            \State Compute right privacy cost: $c_{e,Right}^{(m)}(\lambda)
            \gets
            \log \mathbb{E}_{K \sim \mathrm{Bin}(\lambda, q)}
            \left[
            \exp\!\left(
            \frac{K^2 + K}{2\sigma^2} \Delta_e^{(m)}
            \right)
            \right]$

            \State Set per-sample cost:
            \[
            \hat c_e^{(m)}(\lambda)
            \gets
            \max\{c_{e,Left}^{(m)}(\lambda), c_{e,Right}^{(m)}(\lambda)\}
            \]
        \EndFor

        \State Compute high-probability upper bound: $\hat c_e(\lambda)
        \gets
        \mathrm{UCB}_{1-\gamma}
        \big(
        \{\hat c_e^{(m)}(\lambda)\}_{m=1}^M
        \big)$

        \State Accumulate privacy cost:
        \[
        C_{\mathrm{tot}}(\lambda)
        \gets
        C_{\mathrm{tot}}(\lambda) + \hat c_e(\lambda)
        \]

    \EndFor
\EndFor

\vspace{0.3em}
\Statex \textcolor{blue}{// Final Bayesian Differential Privacy Guarantee}

\For{$\lambda \in \Lambda$}
    \State Compute privacy bound: $\epsilon_\mu(\lambda)
    \gets
    \frac{C_{\mathrm{tot}}(\lambda) - \log \beta}{\lambda}$
\EndFor

\State Select optimal privacy level: $\epsilon_\mu \gets \min_{\lambda \in \Lambda} \epsilon_\mu(\lambda)$

\State \Return $\theta_T$, $(\epsilon_\mu, \delta_\mu = \beta + \gamma)$

\end{algorithmic}
*Note: In practice, this bound $\mathrm{UCB}_{1-\gamma}$ can be obtained using a Bayesian posterior quantile. 

\end{algorithm}

In the remainder of this paper, classical $(\epsilon,\delta)$-DP is used as a baseline, while our proposed framework employs Bayesian differential privacy with parameters $(\epsilon_\mu,\delta_\mu)$.

\subsection{Federated Survival Learning with Bayesian Differential Privacy}
\label{subsec:bayesian_dp}
In this section, we adapt federated learning with Bayesian Differential Privacy
\citep{triastcyn2020bayesian} to a credit risk survival analysis framework,
providing tighter, data-aware privacy bounds than standard federated learning
with classical differential privacy.
Algorithm~\ref{alg:bayesian_dp_federated} summarizes the complete training
procedure for federated survival learning under Bayesian Differential Privacy,
highlighting the interaction between local survival optimization,
posterior-based privacy accounting, and global model aggregation. Bayesian privacy guarantees are enforced at the client level during local
optimization, while federated aggregation constitutes post-processing and does
not incur additional privacy loss.

\section{Experimental Setup}
\label{sec:EXPERIMENT_SETUP}

This section describes the experiments for our proposed FSL credit risk estimation model. We outline the datasets, the objective function and target definition, the performance metrics, and the training and validation setup. 

\subsection{Dataset Description}
We evaluate FSL-BDP on three real-world credit datasets spanning different lending contexts, geographies, and default dynamics. Table~\ref{tab:dataset_stats} summarizes key characteristics of the used datasets. These datasets present realistic decision support challenges spanning different lending contexts. LendingClub involves consumer credit decisions requiring rapid, automated assessment at scale. SBA involves small business lending with longer horizons, government guarantees, and different risk dynamics. Bondora spans multiple European countries with heterogeneous regulatory environments and borrower populations. Each context demands different decision support considerations from real-time scoring to long-horizon provisioning, making this a comprehensive evaluation of FSL-BDP's practical applicability.

\begin{itemize}
    \item \textbf{LendingClub}: The first dataset is drawn from LendingClub, a major U.S. peer-to-peer lending platform widely used in credit risk research~\cite{zhang2020credit,serrano2016use,jagtiani2019roles,lyocsa2022default,croux2020important,papouskova2019two}. The dataset include approximately 150 features spanning application information, credit bureau summaries, borrower demographics, and repayment behavior. The event indicator is derived from the \texttt{loan\_status} field confirming charge-off or default, with default defined as occurring within a 24-month observation window from loan origination.
    \item \textbf{SBA Dataset}: The second dataset is drawn from the U.S. Small Business Administration (SBA) loan program, which comprises government-guaranteed loans to small businesses and is widely used in credit risk research~\cite{li2018should,glennon2005measuring}. The data include loan terms, guarantee characteristics, borrower attributes (industry, firm age, employment), and macroeconomic indicators. The event indicator is defined using the \texttt{ChgOffDate} field, with default observed within a 48-month window from origination. Given the long tenure of SBA loans, the extended observation window ensures a sufficient event rate for survival modeling.
    \item \textbf{Bondora}: The third dataset is drawn from Bondora, a leading European peer-to-peer lending platform~\citep{bondora_site,lyocsa2022default,mondal2023predicting,domotor2023peer,bone2024improving}. The dataset contains borrower demographic characteristics, financial attributes, and loan-level transactional information. The event indicator is constructed from the \texttt{ActiveLateLastPaymentCategory} field, where an event is defined as reaching 90+ days past due within a 24-month observation window from loan origination.

\end{itemize}
Standard preprocessing steps were applied: date features were transformed into days-from-reference, columns prone to data leakage (e.g., \texttt{funded\_amnt}, \texttt{int\_rate}) were removed, missing values were imputed with zeros, categorical variables were one-hot encoded with infrequent categories grouped, and standardization was applied to the training set only to prevent leakage.

\begin{table}[H]
\centering
\label{tab:dataset_stats}
\caption{Dataset Statistics}
\begin{tabular}{@{}lccc@{}}
\toprule
\textbf{Characteristic} & \textbf{LendingClub} & \textbf{SBA} & \textbf{Bondora} \\
\midrule
Total observations & 1,660,791 & 753,258 & 218,559 \\
Number of features & 59 & 33 & 37 \\
Default rate & 12.8\% & 4.5\% & 32.5\% \\
Observation window & 24 months & 48 months & 24 months \\
\midrule
\multicolumn{4}{l}{\textit{Federated Partitioning}} \\
\midrule
Number of clients ($K$) & 27 states & 32 states & 3 countries \\
Min client size & 20,327 & 8,244 & 25,916 \\
Max client size & 229,389 & 114,088 & 124,349 \\
Mean client size & 54,938 & 26,175 & 72,761 \\
\midrule
\multicolumn{4}{l}{\textit{Data Splits}} \\
\midrule
Train/Test/OOT ('K) & 978/244/438 & 502/125/126 & 127/32/59 \\
Train/Test window & 2013-2016 & 1985-2006 & 2013-2018 \\
OOT window & 2017 & 2007-2011 & 2019- Sep 2022 \\
\bottomrule
\end{tabular}
\end{table}

\begin{figure}[htbp]
  \centering
  \includegraphics[width=0.7\linewidth]{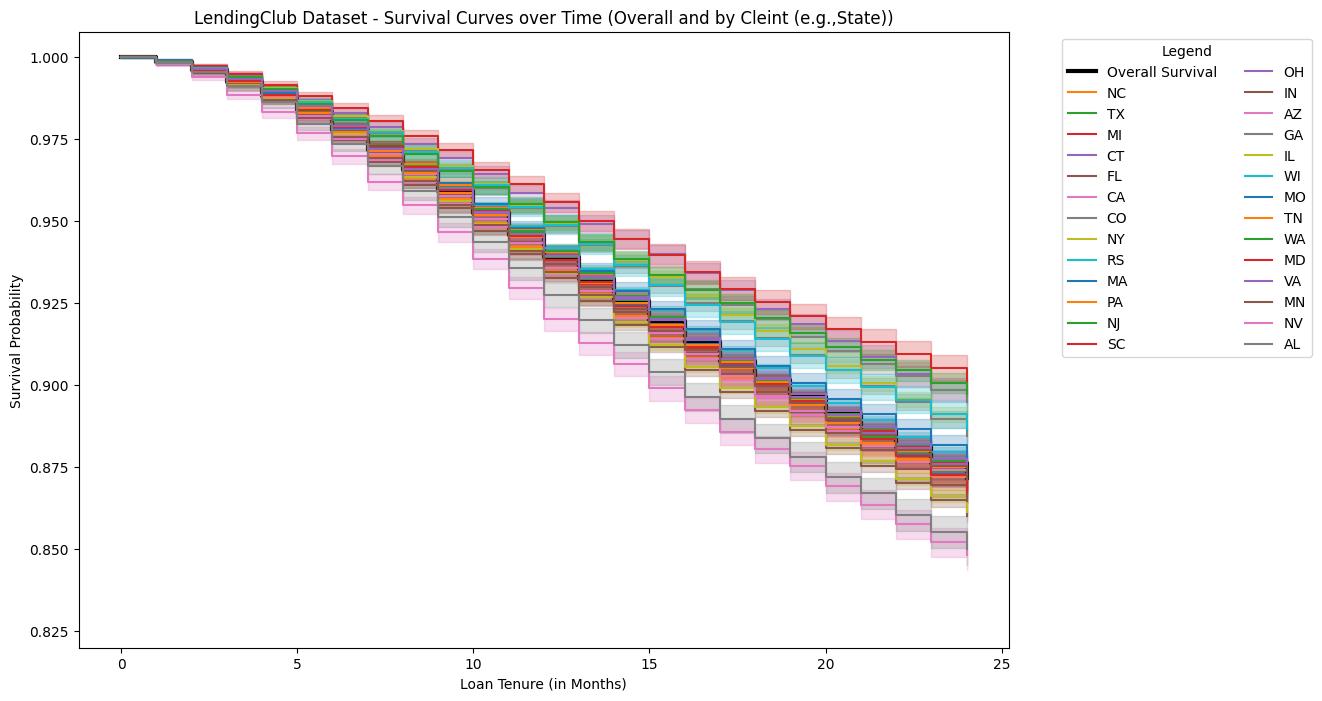}
  \caption{Survival Curve – LendingClub P2P Dataset}
  \label{fig:lc_curve}
\end{figure}

\begin{figure}[htbp]
  \centering
  \includegraphics[width=0.7\linewidth]{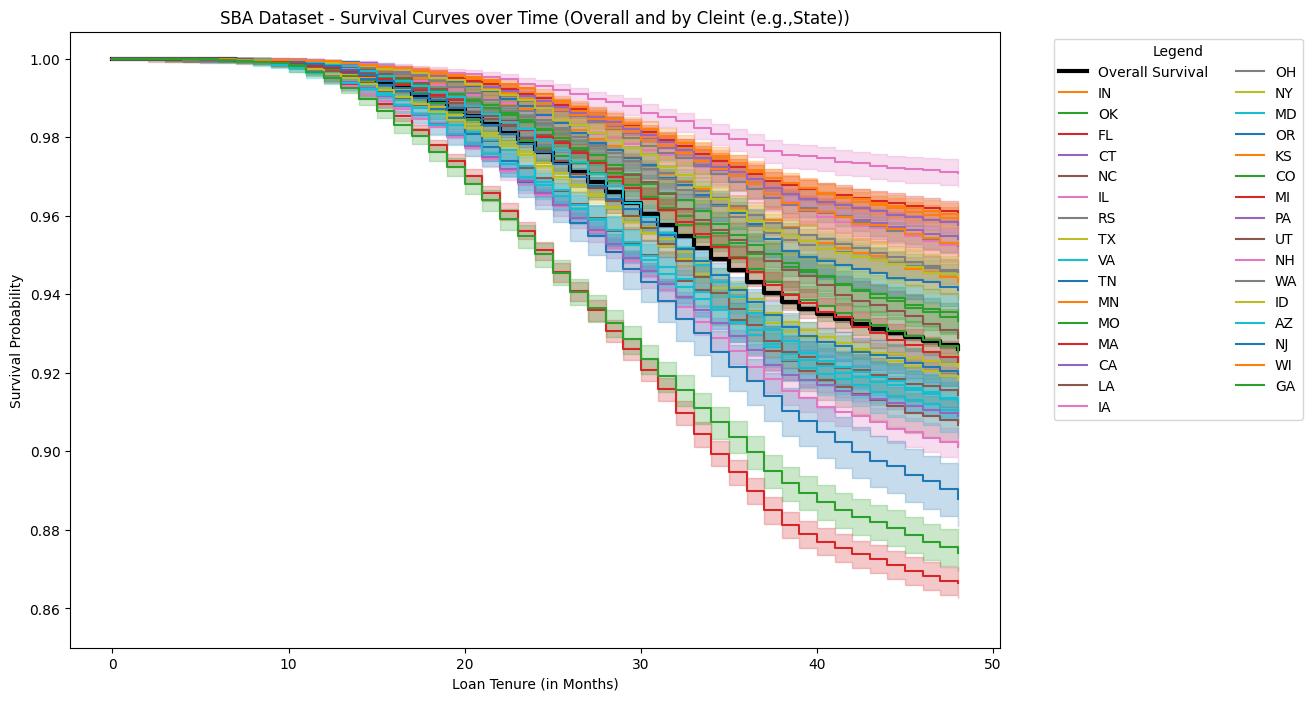}
  \caption{Survival Curve – SBA Loan Dataset}
  \label{fig:sba_curve}
\end{figure}

\begin{figure}[htbp]
  \centering
  \includegraphics[width=0.7\linewidth]{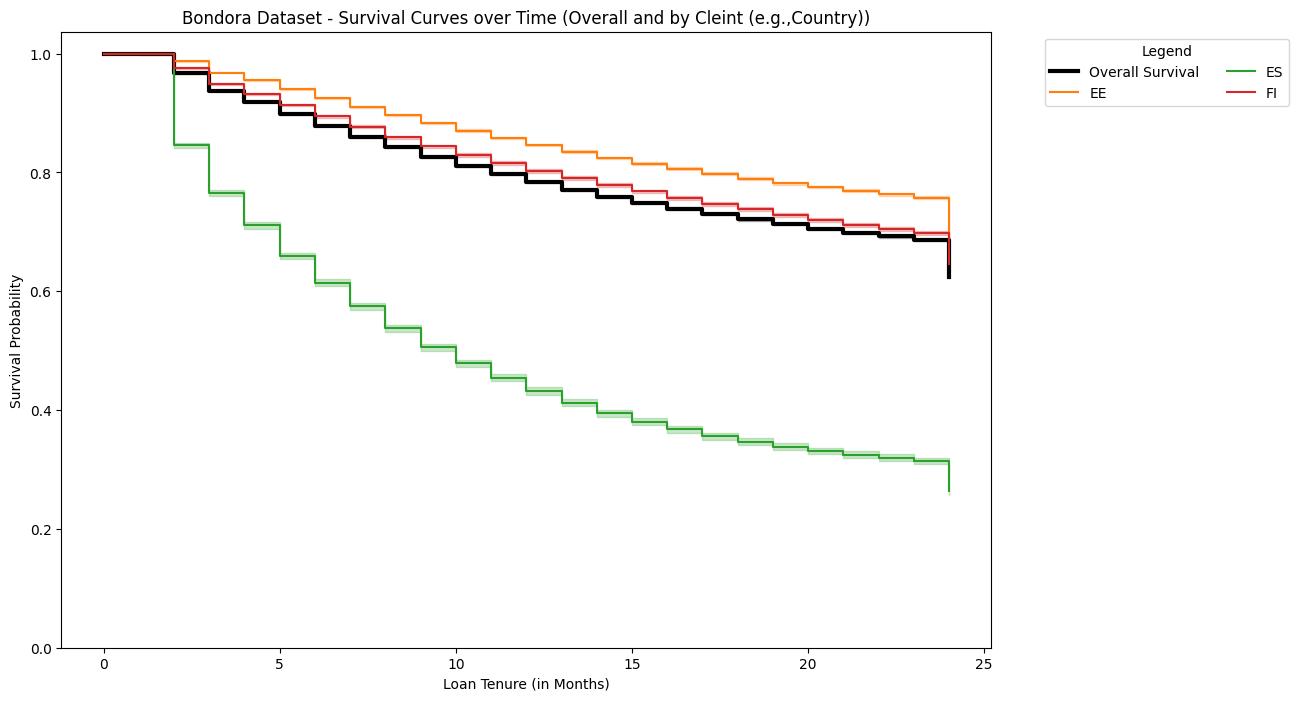}
  \caption{Survival Curve – Bondora Dataset}
  \label{fig:bondora_curve}
\end{figure}

\subsection{Performance Evaluation Metrics}
We evaluated each model's performance using two established survival analysis metrics: the concordance index for discrimination ability and the integrated Brier score for prediction accuracy. Both metrics are computed on held-out test sets to ensure unbiased performance estimation.

\paragraph{Concordance Index (C-index)}
\label{sec:cindex}

The concordance index measures the model's ability to correctly rank subjects according to their survival risk \cite{steck2007ranking}. For time-dependent predictions, we employ the time-dependent C-index \cite{antolini2005time}, which evaluates discrimination at specific time horizons while accounting for censoring. At evaluation time $t^*$, consider all admissible pairs $(i,j)$ where subject $i$ experienced an event before $t^*$ ($T_i \leq t^*, \delta_i = 1$) and subject $j$ was still at risk ($T_j > T_i$). The pair is concordant if the model correctly predicts lower survival probability for subject $i$:
\begin{equation}
\label{eq:cindex}
C(t^*) = \frac{\sum_{i,j} \mathbbm{1}(T_i \leq t^*, \delta_i = 1, T_j > T_i) \cdot \mathbbm{1}(\hat{S}(T_i \mid \mathbf{x}_i) < \hat{S}(T_i \mid \mathbf{x}_j))}{\sum_{i,j} \mathbbm{1}(T_i \leq t^*, \delta_i = 1, T_j > T_i)},
\end{equation}
where $\hat{S}(t \mid \mathbf{x})$ is the predicted survival probability at time $t$ for covariates $\mathbf{x}$, $T_i$ is the observed time, $\delta_i$ is the event indicator, and $\mathbbm{1}(\cdot)$ denotes the indicator function. The C-index ranges from 0.5 (random discrimination) to 1.0 (perfect discrimination). Values above 0.7 indicate acceptable discrimination; values above 0.8 indicate strong predictive ability. We report the mean C-index across evaluation time points $\mathcal{T}_{\text{eval}}$:
\begin{equation}
\label{eq:cindex_mean}
\bar{C} = \frac{1}{|\mathcal{T}_{\text{eval}}|} \sum_{t^* \in \mathcal{T}_{\text{eval}}} C(t^*).
\end{equation}

\paragraph{Integrated Brier Score (IBS)}
\label{sec:ibs}

The Brier score quantifies prediction accuracy by measuring the mean squared error between predicted survival probabilities and observed outcomes \cite{graf1999assessment}. Unlike the C-index, which only evaluates ranking, the Brier score assesses both discrimination and calibration. For survival data with censoring, we use the inverse probability of censoring weighting (IPCW) to obtain unbiased estimates. At evaluation time $t^*$, the censoring-adjusted Brier score is:
\begin{equation}
\label{eq:brier}
\text{BS}(t^*) = \frac{1}{n} \sum_{i=1}^{n} \left[ \frac{\mathbbm{1}(T_i \leq t^*, \delta_i = 1)}{\hat{G}(T_i)} \hat{S}(t^* \mid \mathbf{x}_i)^2 + \frac{\mathbbm{1}(T_i > t^*)}{\hat{G}(t^*)} (1 - \hat{S}(t^* \mid \mathbf{x}_i))^2 \right],
\end{equation}
where $\hat{G}(t)$ is the Kaplan-Meier estimate of the censoring distribution, computed on the training data with reversed event indicators. The integrated Brier score aggregates performance across the evaluation time range $[t_{\min}, t_{\max}]$:
\begin{equation}
\label{eq:ibs}
\text{IBS} = \frac{1}{t_{\max} - t_{\min}} \int_{t_{\min}}^{t_{\max}} \text{BS}(t) \, dt.
\end{equation}

In practice, we approximate this integral using the trapezoidal rule over discrete evaluation times $\mathcal{T}_{\text{eval}} = \{t_1, \ldots, t_m\}$:
\begin{equation}
\label{eq:ibs_discrete}
\text{IBS} \approx \frac{1}{t_m - t_1} \sum_{k=1}^{m-1} \frac{\text{BS}(t_k) + \text{BS}(t_{k+1})}{2} \cdot (t_{k+1} - t_k).
\end{equation}

The IBS ranges from 0 (perfect prediction) to 1, with lower values indicating better overall accuracy. The IBS provides a comprehensive performance measure as it penalizes both poor discrimination and miscalibration.

\begin{figure}
    \centering  \includegraphics[width=1.0\linewidth]{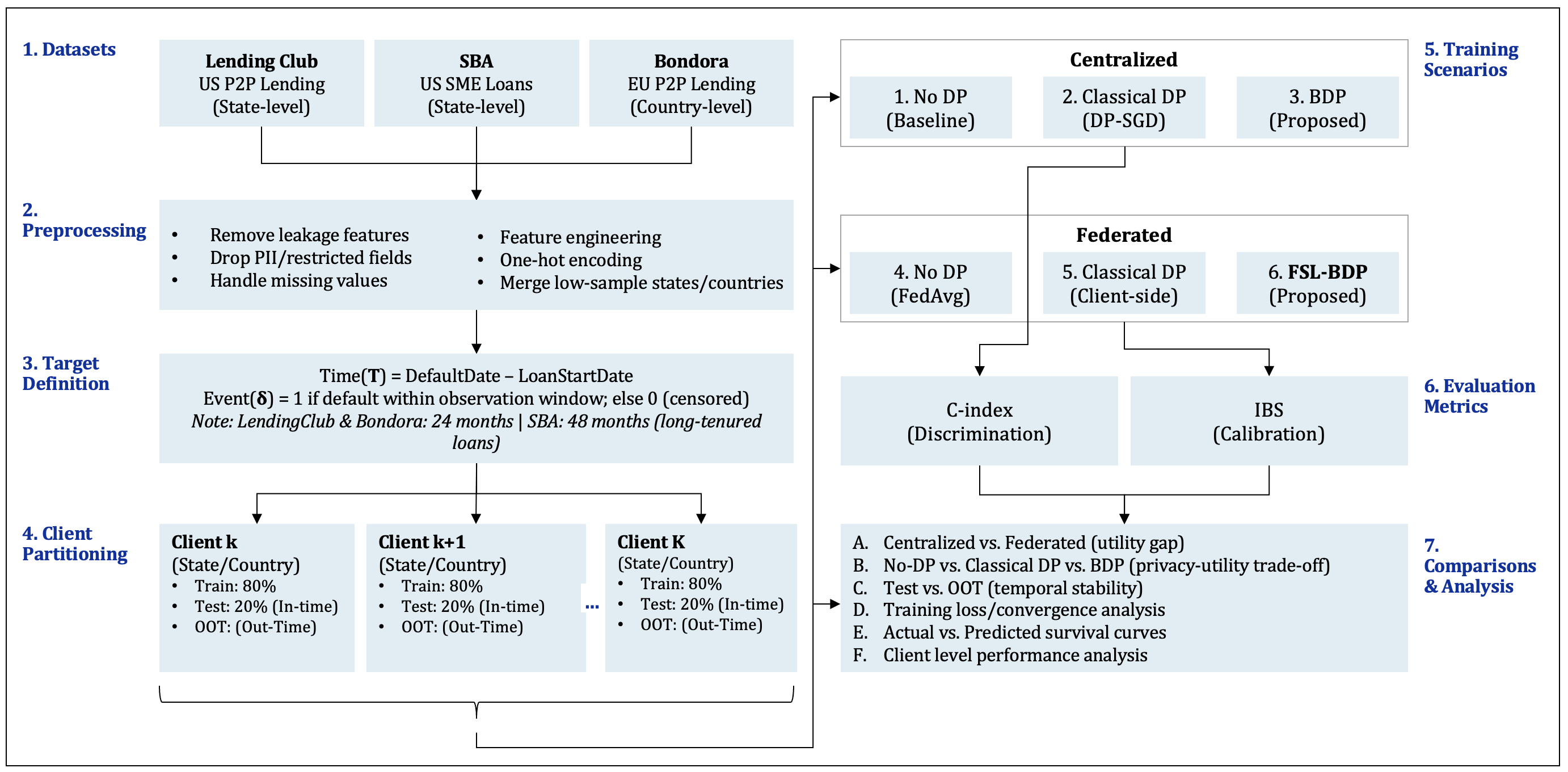}
    \caption{Experiment design overview}
    \label{fig:experiment_design}
\end{figure}

\subsection{Implementation Details}

\label{sec:Implementation_Details}
We implemented FSL-BDP in PyTorch 2.0 to simulate a federated scenario with one central server and $K$ participating clients, executed on NVIDIA T4 GPUs. Unlike prior federated credit scoring studies that rely on artificial client partitions through random sampling~\cite{shingi2020federated, wang2024novel, li2024dynamic}, we partition datasets according to natural geographic boundaries: LendingClub and SBA by U.S. states (27 and 32 clients), Bondora by European countries (3 clients). This yields organically heterogeneous distributions reflecting real cross-regional variation in default behavior, as shown in Figures~\ref{fig:lc_curve},~\ref{fig:sba_curve}, and~\ref{fig:bondora_curve}. Following industry practice, each client's data is split into training (80\%), in-time testing (20\%), and out-of-time validation with data samples outside training window to stress test model robustness.

The survival model employs a six-layer feed-forward network (Input$\rightarrow$128$\rightarrow$64$\rightarrow$64$\rightarrow$32$\rightarrow$32$\rightarrow$$T$) with SELU activations and sigmoid output to ensure hazard probabilities $h_j(x) \in (0,1)$ across $T$ discrete time intervals. Training uses Adam optimizer ($\eta = 0.001$), batch size $B = 32$, and the negative log-likelihood loss for discrete-time survival (Equation~\ref{eq:vectorized_loglik}). The federated protocol, illustrated in Figure~\ref{fig:experiment_design}, runs $R = 10$ communication rounds with full client participation, where each client performs $E = 5$ local epochs before uploading privatized updates for weighted FedAvg aggregation.

For differential privacy, we set gradient clipping threshold $C = 1.0$ and calibrate noise multiplier $\sigma$ to achieve target budgets $\varepsilon \in \{0.5, 1.0, 2.0, 10.0\}$ with $\delta = 10^{-5}$. The Bayesian DP mechanism employs leave-one-out gradient estimation with $M = 10$ sub-batches per batch; clients with insufficient data ($n_k < 100$) fall back to analytical DP with closed-form guarantees.

All experiments use fixed random seeds (seed = 42) with deterministic CUDA settings for reproducibility. Results are averaged over 5 independent runs with standard deviations reported in Table~\ref{tab:results}.

\section{Experimental Results and Analysis}
\label{sec:Results_Discussions}
This section presents experimental results for FSL-BDP across three credit datasets. We examine: (i) performance under centralized and federated training with different privacy settings, (ii) convergence behavior, (iii) the model's ability to distinguish early from late defaulters, (iv) calibration of predicted survival curves, and (v) client-level variation in performance.

\subsection{Overall Performance Comparison}
\label{subsec:overall_performance}
We compare discrimination and calibration performance across centralized and federated learning settings under three privacy regimes: no privacy (No-DP), Classical Differential Privacy (DP), and Bayesian Differential Privacy (BDP). Performance is evaluated using the concordance index (C-index) and integrated Brier score (IBS) on both in-time-test(Test) sets and out-of-time (OOT) validation sets, with detailed results summarized in Table~\ref{tab:results}.

\begin{table*}[htbp]
\centering
\caption{Performance Comparison Across Datasets and Testing Scenarios. Results are reported for centralized and federated training under no privacy, classical differential privacy (DP), and Bayesian differential privacy (BDP)}
\label{tab:results}
\small
\begin{tabular}{llcccc}
\toprule
\textbf{Dataset} & \textbf{Scenario} & \textbf{Test CI} & \textbf{Test IBS} & \textbf{OOT CI} & \textbf{OOT IBS} \\
\midrule
\multirow{6}{*}{LendingClub} 
 & Centralized (No DP) & 0.674 $\pm$ 0.009 & 0.055 $\pm$ 0.007 & 0.642 $\pm$ 0.011 & 0.054 $\pm$ 0.008 \\
 & Centralized (Classical DP) & 0.610 $\pm$ 0.011 & 0.063 $\pm$ 0.008 & 0.598 $\pm$ 0.013 & 0.137 $\pm$ 0.018 \\
 & Centralized (BDP) & 0.562 $\pm$ 0.011 & 0.062 $\pm$ 0.008 & 0.563 $\pm$ 0.010 & 0.125 $\pm$ 0.015 \\
 & Federated (No DP) & 0.667 $\pm$ 0.011 & 0.056 $\pm$ 0.007 & 0.644 $\pm$ 0.011 & 0.131 $\pm$ 0.017 \\
 & Federated (Classical DP) & 0.633 $\pm$ 0.013 & 0.063 $\pm$ 0.008 & 0.615 $\pm$ 0.012 & 0.139 $\pm$ 0.018 \\
 & Federated (BDP) & 0.638 $\pm$ 0.012 & 0.059 $\pm$ 0.007 & 0.618 $\pm$ 0.011 & 0.132 $\pm$ 0.017 \\
\midrule
\multirow{6}{*}{Bondora} 
 & Centralized (No DP) & 0.610 $\pm$ 0.037 & 0.139 $\pm$ 0.050 & 0.542 $\pm$ 0.002 & 0.265 $\pm$ 0.073 \\
 & Centralized (Classical DP) & 0.576 $\pm$ 0.031 & 0.175 $\pm$ 0.069 & 0.511 $\pm$ 0.007 & 0.465 $\pm$ 0.055 \\
 & Centralized (BDP) & 0.550 $\pm$ 0.029 & 0.153 $\pm$ 0.056 & 0.505 $\pm$ 0.005 & 0.428 $\pm$ 0.050 \\
 & Federated (No DP) & 0.602 $\pm$ 0.037 & 0.177 $\pm$ 0.111 & 0.529 $\pm$ 0.008 & 0.306 $\pm$ 0.057 \\
 & Federated (Classical DP) & 0.576 $\pm$ 0.019 & 0.210 $\pm$ 0.116 & 0.517 $\pm$ 0.006 & 0.476 $\pm$ 0.030 \\
 & Federated (BDP) & 0.582 $\pm$ 0.021 & 0.176 $\pm$ 0.093 & 0.515 $\pm$ 0.005 & 0.436 $\pm$ 0.023 \\
\midrule
\multirow{6}{*}{SBA} 
 & Centralized (No DP) & 0.748 $\pm$ 0.034 & 0.019 $\pm$ 0.007 & 0.639 $\pm$ 0.047 & 0.089 $\pm$ 0.036 \\
 & Centralized (Classical DP) & 0.696 $\pm$ 0.044 & 0.020 $\pm$ 0.008 & 0.596 $\pm$ 0.056 & 0.096 $\pm$ 0.041 \\
 & Centralized (BDP) & 0.692 $\pm$ 0.036 & 0.020 $\pm$ 0.008 & 0.612 $\pm$ 0.053 & 0.094 $\pm$ 0.039 \\
 & Federated (No DP) & 0.738 $\pm$ 0.041 & 0.019 $\pm$ 0.007 & 0.631 $\pm$ 0.044 & 0.077 $\pm$ 0.033 \\
 & Federated (Classical DP) & 0.699 $\pm$ 0.047 & 0.021 $\pm$ 0.008 & 0.576 $\pm$ 0.057 & 0.099 $\pm$ 0.042 \\
 & Federated (BDP) & 0.704 $\pm$ 0.043 & 0.020 $\pm$ 0.008 & 0.598 $\pm$ 0.056 & 0.095 $\pm$ 0.040 \\
\bottomrule
\end{tabular}
\begin{tablenotes}
\small
\item \textit{Note:} CI = Concordance Index (higher is better); IBS = Integrated Brier Score (lower is better). Metrics reported as mean $\pm$ std. 
\end{tablenotes}
\end{table*}

Three patterns emerge from the results. First, federated learning without privacy constraints performs comparably to centralized training. Across all datasets, the C-index drop from federation is less than 1\% in both test and out-of-time validation. This validates, survival-based models for credit scoring can be trained collaboratively without meaningful loss in discrimination, even with heterogeneous client data. Second, under privacy constraints, Bayesian DP outperforms Classical DP in federated settings. Federated BDP achieves higher C-index and lower IBS than federated Classical DP across all datasets. However, the ranking reverses in centralized training, where Classical DP performs better. Privacy mechanism rankings from centralized benchmarks do not transfer reliably to federated deployments. Third, Bayesian DP shows better temporal stability. In out-of-time validation, federated BDP degrades less than Classical DP across all datasets. For credit risk applications, stability under distribution shift affects provisioning and capital decisions.

\subsection{Convergence Analysis and Training Dynamics}
\label{subsec:convergence}
\begin{figure}[htbp]
\centering
\includegraphics[width=\textwidth]{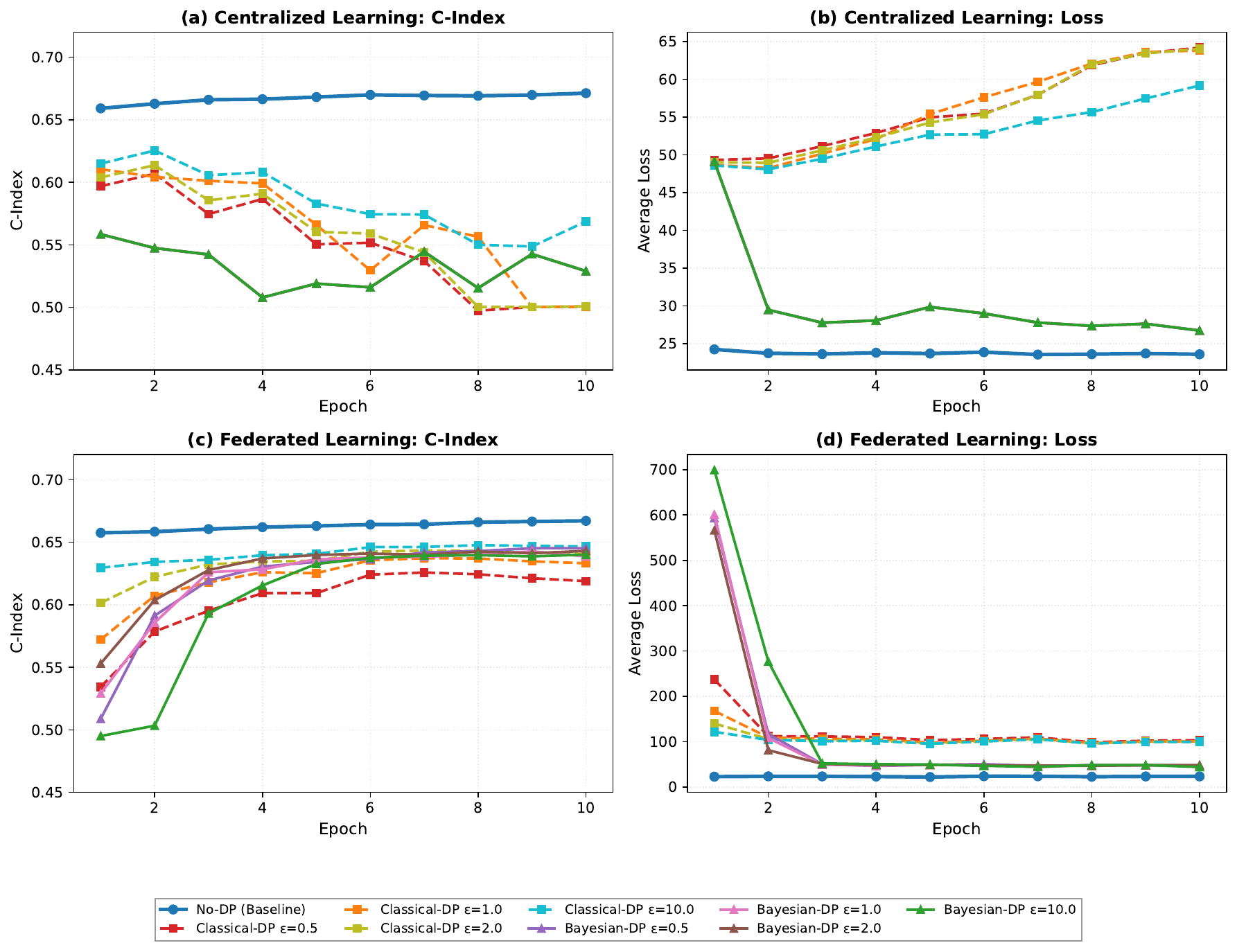}
\caption{Convergence analysis comparing privacy mechanisms across learning paradigms on LendingClub data. 
Panels (a) and (b) show centralized learning; panels (c) and (d) show federated learning. 
The No-DP baseline (blue circles) achieves the highest C-Index but requires data centralization. 
Classical DP (dashed lines, squares) exhibits degraded convergence and increasing loss, particularly at strict privacy budgets ($\varepsilon \leq 1.0$). 
Bayesian DP (solid lines, triangles) demonstrates superior stability with lower loss and competitive discrimination performance, suggesting the proposed FSL-BDP framework's effectiveness in preserving model utility under Bayesian privacy guarantees.}
\label{fig:convergence_combined}
\end{figure}

We analyze training dynamics to understand how different privacy mechanisms affect optimization stability in centralized and federated survival learning. Convergence behavior is evaluated using training loss and C-index trajectories over multiple epochs on the LendingClub dataset, with results illustrated in figure~\ref{fig:convergence_combined}.

Under centralized training, Classical DP shows unstable convergence at strict privacy budgets. Training loss increases over epochs, and discrimination approaches random performance. Fixed worst-case noise appears to disrupt gradient directions enough to prevent effective optimization. In comparison, Bayesian DP shows stable loss reduction and maintains higher discrimination under the same privacy constraints, indicating that data-dependent noise calibration better preserves informative gradients during training.

Federated learning further alters these dynamics. Under federation, both privacy mechanisms demonstrate improved stability, with Bayesian DP benefiting the most. At comparable privacy levels, federated Bayesian DP converges faster, reaches lower training loss, and retains a larger fraction of non-private performance than its centralized counterpart. This pattern is consistent with the presence of implicit regularization induced by client-level gradient aggregation, which moderates noise effects across heterogeneous data partitions. Overall, the results indicate that convergence properties observed under centralized training do not directly translate to federated settings, and that federated architectures can materially improve the stability of privacy-preserving survival learning.

\begin{figure}[H]
\centering
\includegraphics[width=1.0\textwidth, keepaspectratio]{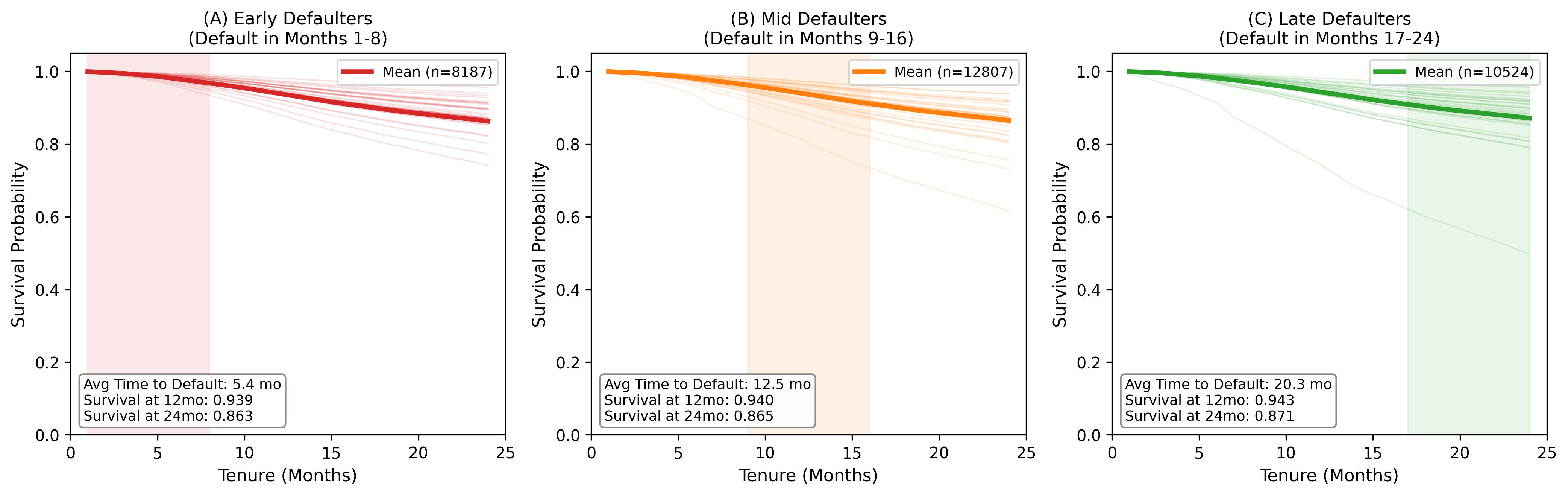}
\caption{Survival Analysis Captures Heterogeneous Default Timing}
\label{fig:survival_curves_v3}
\end{figure}

\subsection{Distinguishing Early and Late Defaulters}
\label{subsec:early_late_defaulters}

Figure~\ref{fig:survival_curves_v3} shows that the survival formulation captures heterogeneity in default timing not observable under binary classification. Survival curves differ across borrower segments, with distinct temporal trajectories.Early defaulters show steep declines in survival probability, with mean predicted default at 5.4 months. Late defaulters maintain high survival probability longer, with mean predicted default at 20.3 months. Binary models would label both groups similarly, but their economic implications differ: early defaults occur before substantial principal repayment, resulting in higher losses. 

This temporal resolution has direct implications for credit risk management and regulatory reporting. The survival framework enables month-by-month hazard estimates that support IFRS~9 lifetime expected credit loss calculations: $\text{ECL} = \sum_{t=1}^{T} h_t(x) \cdot \text{LGD}(t) \cdot \text{EAD}(t)$, where both exposure and loss severity vary over time. In contrast, conventional classification models produce a single risk score without timing information and therefore cannot support loss provisioning or monitoring processes that depend on the evolution of risk over the loan life. These results illustrate how survival-based modeling provides risk signals that are both economically interpretable and operationally relevant.

\begin{figure}[H]
\centering
\includegraphics[width=\textwidth]{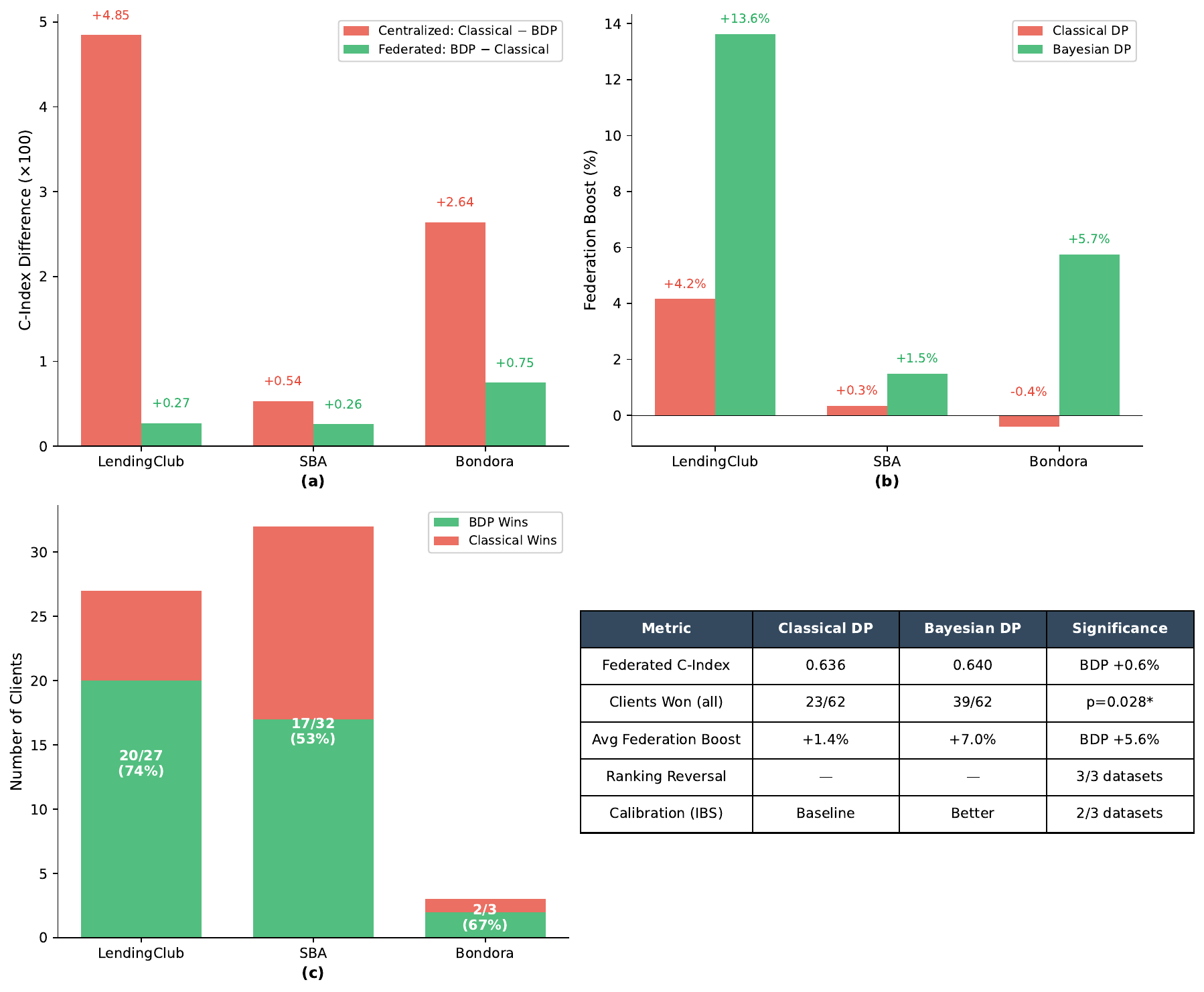}
\caption{Cross-dataset client-level analysis (LendingClub: 27 clients, SBA: 32 clients, Bondora: 3 clients). 
(a)~Ranking reversal is consistent across all three datasets: Classical DP outperforms BDP in centralized training (red bars), but BDP outperforms Classical DP in federated settings (green bars). 
(b)~BDP benefits substantially more from federation (+7.0\% average) than Classical DP (+1.4\%), with Classical DP showing negative federation effect on Bondora. 
(c)~Client-level win rates confirm BDP's advantage: 20/27 (74\%) on LendingClub, 17/32 (53\%) on SBA, 2/3 (67\%) on Bondora. 
(d)~Summary statistics show BDP wins 39/62 clients overall ($p=0.028$), with 5$\times$ higher federation boost.}
\label{fig:client_analysis}
\end{figure}

\begin{figure}[htbp]
\centering
\includegraphics[width=0.90\textwidth, keepaspectratio]{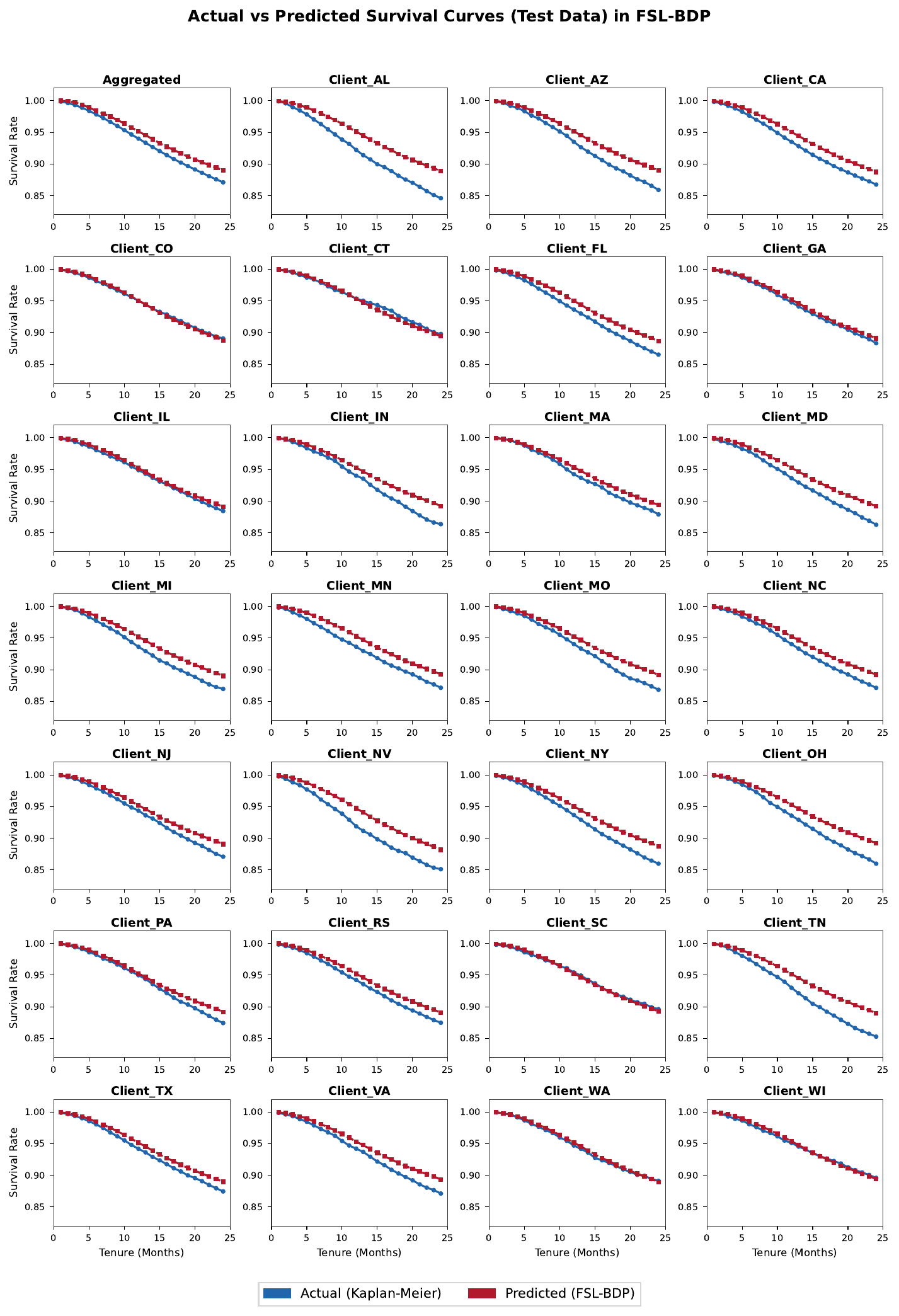}
\caption{Actual vs Predicted Survival Curves on LendingClub Test Data.}
\label{fig:survival_curves_v2}
\end{figure}

\subsection{Model Calibration: Actual vs. Predicted Survival Curves}
\label{subsec:calibration}

We evaluate the calibration of FSL-BDP by comparing predicted survival curves against non-parametric Kaplan-Meier estimates on the LendingClub test data, both at the aggregated level and across 27 state-level clients, illustrated in figure~\ref{fig:survival_curves_v2} .

At the aggregate level, the predicted survival curve closely follows the Kapla-Meier estimate over the full 24-month horizon, capturing the gradual decline in survival probability without systematic over- or underestimation. This indicates that the model provides well-calibrated survival probabilities at the portfolio level. More importantly, the client-level panels show that FSL-BDP adapts effectively to heterogeneous regional risk profiles while learning a single global model. States with higher observed default risk, such as Nevada and Arizona, exhibit steeper survival declines, whereas lower-risk states, including North Carolina and Washington, display flatter trajectories. In each case, the predicted curves remain within the Kaplan--Meier confidence bands, indicating consistent calibration across clients with differing risk characteristics.

Quantitative calibration results support these visual findings. Bayesian DP achieves an integrated Brier score of 0.059, improving upon Classical DP (0.063) and approaching the non-private baseline (0.056). This suggests that Bayesian privacy preserves probability estimation accuracy more effectively than Classical DP under federated training. Reliable calibration is essential in regulated credit risk applications, where Basel III/IV and IFRS~9 require accurate probability estimates for capital adequacy and lifetime loss provisioning. By maintaining both discrimination and calibration under Bayesian privacy guarantees, FSL-BDP addresses a key practical constraint in deploying privacy-preserving models within regulated financial environments.

\subsection{Client-Level Performance Analysis}
\label{subsec:client_analysis}

Figure~\ref{fig:client_analysis} presents client-level performance comparisons across all three datasets and highlights a consistent empirical pattern. While Classical DP achieves higher performance than Bayesian DP under centralized training, this ordering reverses under federated learning across LendingClub, SBA, and Bondora.

The magnitude of this reversal is substantial. When transitioning from centralized to federated training, Bayesian DP exhibits a markedly larger performance gain than Classical DP. Averaged across datasets, Bayesian DP improves by 7.0\%, whereas Classical DP improves by only 1.4\%. This contrast is most pronounced for LendingClub and Bondora, where Classical DP shows limited or negative change under federation, while Bayesian DP consistently improves.

Client-level comparisons further confirm the robustness of this pattern. Across all datasets, Bayesian DP outperforms Classical DP in 39 of 62 clients. Dataset-specific win rates are 74\% for LendingClub, 53\% for SBA, and 67\% for Bondora. A binomial test indicates that this difference is statistically significant ($p = 0.028$), suggesting that the observed advantage of Bayesian DP under federated training is unlikely to be due to random variation.

\section{Discussion}
\label{sec:Discussion}
In this section, we discuss our methodological contributions, practical implications, and, finally, limitations and further research directions.

\subsection{Methodological Contributions}
This study introduces FSL-BDP, a framework for privacy-preserving federated survival learning in credit risk. It enables multiple institutions to train models collaboratively without sharing raw data. Unlike prior federated approaches that use binary classification, FSL-BDP models time-to-default, distinguishing early from late defaulters. This temporal information is relevant for credit monitoring and lifetime risk assessment.

To our knowledge, this is the first application of Bayesian differential privacy to federated survival analysis. The proposed leave-one-out gradient estimation with data-dependent privacy accounting delivers practical $(\varepsilon,\delta)$ guarantees than classical worst-case differential privacy while preserving calibration and predictive utility under realistic constraints. A key finding is that privacy mechanism rankings reverse under federation: Classical DP performs better centralized, but Bayesian DP performs better federated. This suggests that privacy mechanisms should be evaluated in the intended deployment architecture, not inferred from centralized benchmarks. The insight applies broadly to privacy-preserving distributed learning beyond credit risk modeling.

\subsection{Practical Implications}
The framework addresses a key constraint in credit risk management: institutions value cross-lender learning but face regulatory, competitive, and reputational barriers to data sharing. FSL-BDP enables collaborative model development without exposing raw borrower data. By limiting information in gradient exchanges, the framework reduces exposure to membership and attribute inference attacks common in federated settings. Our results show that federation without privacy constraints retains over 99\% of centralized performance. Under privacy constraints, Bayesian DP improves 7.0\% on average when moving from centralized to federated training, compared to only 1.4\% for Classical DP. 

Smaller clients benefit proportionally more from federation, suggesting the framework is particularly valuable for regional lenders with limited individual portfolios. From a regulatory standpoint, calibration results indicate that probability estimates remain valid for regulatory reporting under strict privacy budgets. The temporal granularity supports Basel capital adequacy calculations alongside provisioning requirements, providing risk managers with outputs aligned to existing compliance workflows.

\subsection{Limitations and Further Research}
This study has limitations that suggest directions for future work. First, we assume full client participation in every communication round. Production deployments face partial participation, client dropout, and asynchronous updates, particularly across institutions with heterogeneous infrastructure. How these dynamics interact with Bayesian privacy accounting remains unexplored. Second, out-of-time performance degrades in some datasets, reflecting known challenges in maintaining model accuracy under distribution shift. Continuous monitoring and periodic retraining are standard practice in centralized credit modeling \cite{amed2025pdx}, but extending these to federated settings is difficult when privacy budgets constrain update frequency. Developing federated monitoring, retraining, and budget management mechanisms within a federated machine learning operations framework is therefore an important research direction. Third, the framework assumes horizontal federation with a shared feature space. Many credit risk applications involve vertically partitioned data, with complementary features held by banks, e-commerce platforms, or telecom providers. Extending FSL-BDP to vertical or hybrid federation would require more complex feature alignment and secure aggregation mechanisms while maintaining computational efficiency for real-time decisioning.

Beyond technical extensions, our findings raise open questions at the intersection of analytics, governance, and economics. How should ownership, accountability, and auditability be defined for models trained collaboratively across institutional and regulatory boundaries? What are the cost-benefit tradeoffs between federated and centralized architectures as portfolio size and organizational complexity scale? Under what conditions does federated aggregation systematically strengthen or weaken different privacy mechanisms? Addressing these questions is essential for translating privacy-preserving federated learning from experimental settings into production decision support systems.

\section{Conclusion}
\label{sec:Conclusion}

This study addresses the challenge of building privacy-preserving decision support systems that model time-to-default across multiple institutions while protecting borrower data through Bayesian differential privacy. The results show that federated survival learning retains near non-private performance under strict privacy budgets, enabling collaborative training without exposing sensitive data. Unlike binary classification, the survival formulation provides temporal risk estimates that directly support operational decisions: loan approval based on expected default timing, risk-adjusted pricing, and IFRS 9 lifetime loss provisioning. A key finding for decision support system design is that privacy mechanism performance depends on deployment architecture. Bayesian DP outperforms Classical DP under federation, reversing rankings observed in centralized settings. Practitioners building privacy-preserving decision support systems should therefore evaluate mechanisms in their intended deployment environment rather than rely on centralized benchmarks. This insight generalizes to other multi-institutional contexts such as fraud detection, healthcare analytics, and supply chain risk, where data sovereignty constraints preclude centralization but collaborative learning remains valuable.

\newpage
\bibliographystyle{unsrt}  
\bibliography{aipsamp}

\end{document}